\documentclass{article}

\usepackage{subfigure}
\usepackage{amssymb}
\usepackage{graphicx}
\usepackage{commath}
\usepackage[ansinew]{inputenc}
\usepackage{graphicx,color}
\usepackage{url}
\usepackage{latexsym,enumerate}
\usepackage{booktabs}
\usepackage{framed}
\usepackage{subfigure}
\usepackage{multirow} 
\usepackage{color}
\usepackage{colortbl}
\usepackage{algorithm}
\usepackage{algorithmic}

\definecolor{gray1}{rgb}{0.8,0.8,0.8}
\definecolor{gray2}{rgb}{0.95,0.95,0.95}

\newcommand{\RE}{\,{\rm Re}}

\newcommand{\argmin}{\mathop{\rm argmin}}
\newcommand{\Shrink}{\mathop{\rm Shrink}}

\newcommand{\CST}{\mathop{\rm CST}}

\begin{document}

\date{}

\title{Global Variational Method for Fingerprint Segmentation by Three-part Decomposition}

\author{D.H. Thai\thanks{Institute for Mathematical Stochastics, 
University of Goettingen,
Goldschmidtstr. 7, 37077 G\"ottingen, Germany.
Email: duy.thai@math.uni-goettingen.de and gottschlich@math.uni-goettingen.de}
\ and C. Gottschlich$^{\ast}$}

\maketitle

\begin{abstract}
Verifying an identity claim by fingerprint recognition is a commonplace experience for millions of people in their daily life, e.g. for unlocking a tablet computer or smartphone. The first processing step after fingerprint image acquisition is segmentation, i.e. dividing a fingerprint image into a foreground region which contains the relevant features for the comparison algorithm, and a background region. We propose a novel segmentation method by global three-part decomposition (G3PD). Based on global variational analysis, the G3PD method decomposes a fingerprint image into cartoon, texture and noise parts. After decomposition, the foreground region is obtained from the non-zero coefficients in the texture image using morphological processing. The segmentation performance of the G3PD method is compared to five state-of-the-art methods on a benchmark which comprises manually marked ground truth segmentation for 10560 images. Performance evaluations show that the G3PD method consistently outperforms existing methods in terms of segmentation accuracy.
\end{abstract}

\section{Introduction} \label{intro}

Fingerprint verification is a widely used authentication method in commercial applications 
and most fingerprint verification systems rely on minutiae for comparing two fingerprints.
Typical steps of fingerprint image processing \cite{Gottschlich2010PhD} 
include segmentation, orientation field estimation \cite{GottschlichMihailescuMunk2009},
image enhancement by contextual filtering \cite{Gottschlich2012,GottschlichSchoenlieb2012}
and minutiae extraction. 
Additionally, many systems include nowadays a software-based liveness detection module 
which can 
e.g. be based on histograms of invariant gradients \cite{GottschlichMarascoYangCukic2014}
as a countermeasure against so-called spoof attacks.
In this paper, we focus on the fingerprint image segmentation step
and we propose a global three-part decomposition (G3PD) method
to achieve an accurate extraction of the foreground region.

\subsection{Global three-part decomposition (G3PD) method}

Our proposed method is based on the paradigm that 
a fingerprint image can be considered as a composition of three components:
texture, homogeneous parts and small scale objects.
The G3PD method
aims to decompose a fingerprint image into the corresponding three parts:

\begin{itemize}
 \item \textbf{Texture image:} By texture we refer to the fact that fingerprint images are highly determined by their oriented patterns 
       which have frequencies only in a specific band in the Fourier spectrum, see~\cite{ThaiHuckemannGottschlich2015}.
 \item \textbf{Cartoon image:} The homogeneous regions correspond to the lower frequency response. 
 \item \textbf{Noise image:} Small scale objects staying in the higher frequency band are considered as noise,
       e.g. black dots with random position and intensity.
\end{itemize}

\begin{figure*}[ht]
\begin{center}
 \subfigure{ \includegraphics[width=0.97\textwidth]{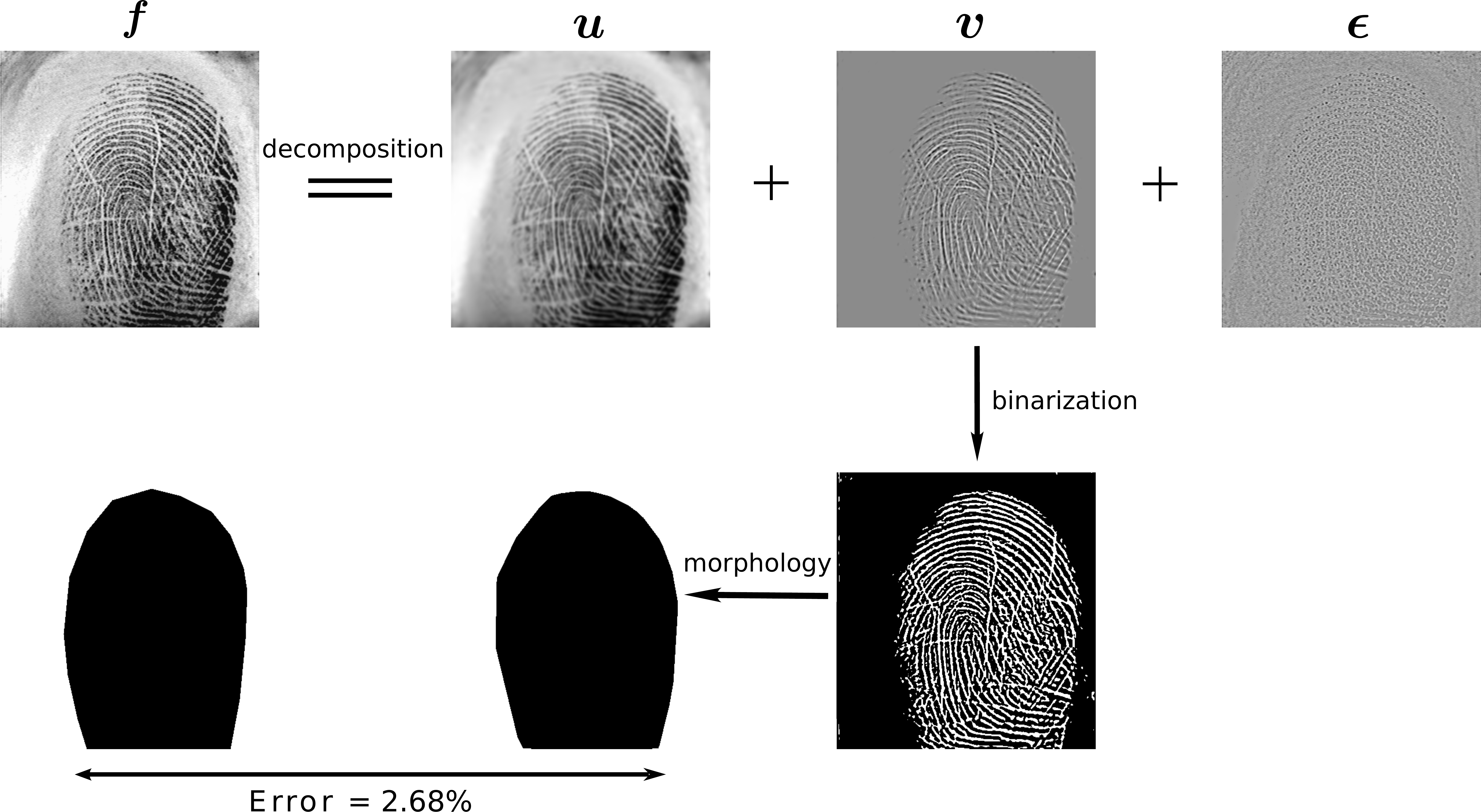} }
 
 \caption{\small Overview over the segmentation by the G3PD method: Firstly, the original image $\boldsymbol f$ 
          is decomposed into
					cartoon image $\boldsymbol u$, texture image $\boldsymbol v$ and noise image $\boldsymbol \epsilon$.
					Secondly, the texture image $\boldsymbol v$ is binarized
					by separating zero from non-zero coefficients
					and the foreground regoin is obtained by morphological operations. 
					In order to evaluate the segmentation performance, 
          the estimated ROI (second row, second column) is compared to manually marked ground truth segmentation (second row, first column).
					Note that the cartoon image $\boldsymbol u$ and noise image $\boldsymbol \epsilon$
					contain also texture parts
					but this choice of parameters leads to a better segmentation performance
					as demonstrated in evaluations on a benchmark with 10560 images, see Section \ref{sec:evaluation}.
         }         
 \label{fig:overviewG3DP}
\end{center}
\end{figure*}

For the purpose of fingerprint segmentation, 
we are only interested in the texture image as a feature for segmentation.
After the decomposition, the cartoon and noise images are ignored. 
Therefore, the decomposition can be considered as a feature extraction step
which has the goal to estimate the best possible texture image for a given input image.
Subsequently, the region of interest (ROI) is obtained by morphological operations 
on the non-zero coefficients in the extracted texture image, see Figure \ref{fig:overviewG3DP}.
In order to achieve these goals, we propose a model for three-part decomposition 
with variational based methods as described below.
The G3PD method follows the same philosophy of texture image extraction
as the Fourier based FDB method~\cite{ThaiHuckemannGottschlich2015},
but regards the problem from a different point of view 
and solves it by a variational approach.

\paragraph{Proposed variational model for G3PD}

Decomposition techniques are at the core of variational methods.
Decomposition is performed by finding the solution of a convex minimisation problem.
Inspired by this idea, we propose a novel model for global three-part decomposition
which has five ingredients: 
\textbf{(1) Cartoon:}
Piecewise constant regions are measured by the anisotropic total variation (TV) norm \cite{GoldsteinOsher2009}. 
\textbf{(2) Texture:}
The sparsity of the texture pattern is measured by the $\ell_1$ norm 
which is well-known to enhance the sparseness of the solution.
\textbf{(3) Texture:}
The smoothness of the texture image is enforced by the $\ell_1$ norm of the curvelet coefficients.
\textbf{(4) Noise:}
Noise is measured by the supremum norm of its curvelet coefficients.
\textbf{(5) Reconstruction constraint:} Finally, the constraint $\boldsymbol f = \boldsymbol u + \boldsymbol v + \boldsymbol \epsilon$ ensures
that the sum of the three component images reconstructs the original image $\boldsymbol f$.
Empirically, we have found that the curvelets capture 
the geometry of fingerprint patterns better than classical wavelets, see Section \ref{sssec:smoothsparse}.

The combination of the decomposition and morphology in our proposed G3PD method 
yields segmentation performance superior to existing segmentation methods.

\paragraph{Performance Evaluation and Comparison to Existing Methods}

We conduct a systematic performance comparison of our proposed G3PD method 
with five state-of-the-art fingerprint segmentation methods.
The segmentation accuracy of all methods is measured 
on a manually marked ground truth database containing 10560 images \cite{ThaiHuckemannGottschlich2015}.
A detailed description of the evaluation benchmark, training and test protocols, and experimental results 
is given in Section~\ref{sec:evaluation}.
The five methods in the comparison are:
a method based on mean and variance of grey level intensities 
and the coherence of gradients as features 
and a neural network as a classifier \cite{BazenGerez2001}, 
a method using Gabor filter bank responses~\cite{ShenKotKoo2001},
a Harris corner response based method~\cite{WuTulyakovGovindaraju2007}, 
an approach using local Fourier analysis~\cite{ChikkerurCartwrightGovindaraju2007}
and the factorized directional bandpass method \cite{ThaiHuckemannGottschlich2015}.

\subsection{Related Work}

With more than hundred methods, we refer the reader to \cite{ThaiHuckemannGottschlich2015} 
for an overview over the literature of fingerprint segmentation methods.
For image segmentation in general, there is a plethora of approaches to solve this problem. 
These are based e.g. on the intensity of pixels \cite{Otsu1979}, \cite{SahooWilkinsYeager1997}, \cite{AlbuquerqueEsquefMelloAlbuquerque2004},
or the evolution of curves for piecewise smooth regions in 
images \cite{ChanVese2001}, \cite{BressonEsedogluVandergheynstThiranOsher2007}, \cite{ChanEsedogluNikolova2012}, \cite{LieLysakerTai2006}. 
Texture segmentation, however, is still an open problem, 
because intensity values are inadequate, e.g. for segmenting fingerprint patterns. 
Methods based on texture descriptors \cite{SagivSochenZeevi2006}, \cite{HouhouThiranBresson2009} 
or finding other meaningful features 
in an observed image for classification have been suggested.

Based on the classical Rudin-Osher-Fatemi (ROF) model \cite{RudinOsherFatemi1992}, 
researchers have proposed numerous approaches in which the regularisation and fidelity terms 
are considered under different functional spaces, such as Besov, Hilbert 
and Banach spaces \cite{AujolGilboaChanOsher2005}, \cite{AujolAubertFeraudChambolle2005}, \cite{AujolGilboa2006}, 
\cite{BuadesLeMorelVese2010}, \cite{VeseOsher2003}, \cite{AubertVese1997}.
Further image denoising approaches use
higher-order derivatives instead of 
total variation
for minimisation \cite{ChanMarquinaMulet2000}, \cite{LysakerLundervoldTai}, 
mean curvature \cite{ZhuChan2012}, 
Euler's elastica \cite{TaiHahnChung2011},
and total variation of the first and second order derivatives \cite{PapafitsorosSchoenlieb2014},
and higher-order PDEs for diffusion solved by directional operator splitting schemes \cite{CalatroniDueringSchoenlieb2014}.
In particular, many signals have sparse or nearly-sparse representations in some transform domain corresponding to 
$\ell_0$ or its regularisation $\ell_1$ \cite{SongZhangHickernell2013}, \cite{YinOsherGoldfarbDarbon2008}, 
\cite{YinGoldfarbOsher2005}, \cite{YangZhang2011}, \cite{CandesWakinBoyd2008}. 
Aujol and Chambolle \cite{AujolChambolle2005} introduced a model for three-part decomposition
which yields a texture image $\boldsymbol v$ using the G-norm. 
An improvement of the G3PD model in comparison to their work
is especially the texture image extraction
by enforcing smoothness and sparsity on the texture image $\boldsymbol v$.
To solve the constrained minimisation problems, various techniques have been suggested 
such as Chambolle's projection 
\cite{Chambolle2004}, splitting Bregman method \cite{GoldsteinOsher2009}, 
iterative shrinkage/thresholding (IST) algorithms \cite{BeckTeboulle2009}, \cite{DaubechiesDefriseMol2004}, \cite{DiasFigueiredo2007}. 
Wu \textit{et al.} \cite{WuTai2010} has proved the equivalence between augmented Lagrangian method 
(ALM), dual methods, and split Bregman iteration. 
We have adopted ALM into our approach to solve the proposed constrained minimisation problem.
\cite{DonohoJohnstone1994}, \cite{ChambolleDeVoreLeeLucier1998} and \cite{ChoiBaraniuk2004} 
show that the shrinkage operator of multiresolution analysis is 
the solution of a variational problem 
when considering signals in Besov space, i.e. $B^{\alpha}_{p,q}$, relating to wavelet coefficients. 
In this paper, we focus on the 
curvelet transform \cite{CandesDonoho2004}, \cite{CandesDemanetDonohoYing2006}, \cite{StarckDonohoCandes2003}, \cite{MaPlonka2010}
which is very suitable for fingerprint patterns with oriented and curved lines.
However, one can easily adopt our approach for 
the shearlet transform \cite{ShearletsBook},
the contourlet transform \cite{DoVetterli2005}, 
or the steerable wavelet transform \cite{UnserVille2010}.

\begin{figure*}[ht]
\begin{center}
    \subfigure[ ]{ \includegraphics[width=0.22\textwidth]{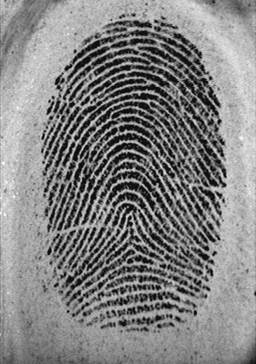} }		
    \subfigure[ ]{ \includegraphics[width=0.22\textwidth]{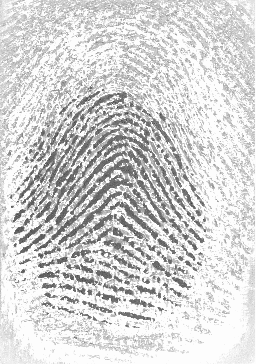} }	
    \subfigure[ ]{ \includegraphics[width=0.22\textwidth]{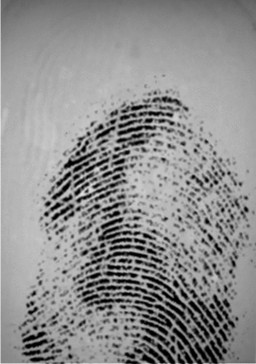} }	
    \caption{Typical difficulties for segmentation encountered in fingerprint analysis.
		(a) Small scale objects and noise on the sensor. (b) Ghost fingerprint. (c) Image with illumination differences. }
    \label{fig:DifficultImage}
\end{center}
\end{figure*}

There are many difficulties relating to the choices of the parameters for decomposition and minimisation steps 
in all aforementioned approaches
which ensure the convergence of the algorithm and 
extract enough texture for segmentation under the various situations, 
such as different illumination, noise, and ghost fingerprints 
(see Figure \ref{fig:DifficultImage} for an illustration). 
To solve these problems is still a challenge in practice.

\subsection{Setup of the paper}

The organisation of the paper is as follows.
In Section \ref{sec:G3PD}, we give a detailed description of the G3PD method 
in two main steps: first, texture image extraction is treated in Section \ref{ssec:decomposition},
followed by morphological operations in Section \ref{ssec:morphology}, see Figure~\ref{fig:overviewG3DP}.
To this end, we introduce the G3PD model
in Section \ref{sssec:model} which defines the objective function 
as a constrained minimisation problem for the decomposition of an image 
into three parts: cartoon, texture and noise images.
Next, in Section \ref{sssec:reformulating} we apply the augmented Langrangian method 
to reformulate the constrained minimisation into an unconstraint one.
Subsequently, this unconstrained minimisation problem is solved 
by the alternating direction method of multipliers (ADMM) in Section \ref{sssec:adm}.
The smoothness and sparsity of the obtained texture image 
as a feature for segmentation is discussed 
in Section \ref{sssec:smoothsparse}.
In Section \ref{ssec:morphology}, we specify how to obtain the ROI 
from the texture image by morphological operations.
In Section \ref{sec:evaluation} we describe the evaluation benchmark, 
the training and test protocols, and experimental results.
Finally, in Section \ref{sec:conclusions} we discuss the results of the evaluation
and we give conclusions.
Additional figures and detailed calculations can be found in \cite{Thai2015PhD}.

\section{The G3PD Method for Fingerprint Segmentation}
\label{sec:G3PD}

This section describes the G3PD method which consists of two main parts: 
in the following Section \ref{ssec:decomposition}, we introduce 
a model for three-part decomposition into cartoon, texture and noise images.
Next, we formalize the constrained minimisation problem
and we discuss the ALM for solving it.
In Section \ref{ssec:morphology}, we utilize the obtained texture image as our feature 
to perform the segmentation by morphological operations.

\subsection{Fingerprint Texture Extraction}
\label{ssec:decomposition}

\subsubsection{The G3PD Model}
\label{sssec:model}
 
As argued before, the fingerprint $\boldsymbol f$ is considered as a composition of a homogeneous region $\boldsymbol u$, 
repeated patterns $\boldsymbol v$ staying in a frequency range in the Fourier domain and corrupted 
by certain random noise~$\boldsymbol \epsilon$. 
Fundamental for our analysis is that we assume that the fingerprint pattern is sparse in the Fourier domain
as the ridge lines form an oscillating signal at essentially one frequency, locally.
A two-dimensional image $f:\Omega \rightarrow \mathbb R_+$
is specified on the lattice with size $N_1 \times N_2$:
$$
\Omega = \left\{\boldsymbol k = (k_1,k_2) \in \big\{ 0, N_1-1 \big\} \times \big\{ 0, N_2-1 \big\} \subset \mathbb N^2 \right\},
$$ 
We assume that
$$
f[\boldsymbol k] ~=~ u[\boldsymbol k] + v[\boldsymbol k] + \epsilon[\boldsymbol k], \qquad \forall \boldsymbol k \in \Omega,
$$
where $\boldsymbol{f}, \boldsymbol{u}, \boldsymbol{v}$ and $\boldsymbol{\epsilon}$ 
are in matrix form, i.e. $\boldsymbol{f} = \big[f[\boldsymbol k]\big]_{\boldsymbol k \in \Omega}$.

The space $B^1_{1,1}$ relating to the $\ell_1$ norm of the wavelet coefficients (cf. \cite{ChambolleDeVoreLeeLucier1998}), 
i.e. $\norm{\boldsymbol v}_{B^1_{1,1}} = \norm{ \mathcal W \{ \boldsymbol v \} }_{\ell_1}$, 
is very suitable to measure the smoothness of the oscillation signals. 
However, due to a set of highly curved lines in the fingerprint patterns,
the $\ell_1$ norm of curvelet coefficients is considered instead to capture their curvature in texture $\boldsymbol v$.
Let $\mathcal C \{\boldsymbol v\} 
       = \big[ \mathcal C_{i,l} \{ \boldsymbol v \} [\boldsymbol k]\big] 
         _{(i,l,\boldsymbol k) \in \mathcal I} 
       $
denote the discrete curvelet transform of $\boldsymbol v$ in $i$ different scales         
and $l$ orientations at positions $\boldsymbol k$ contained in the index set $\mathcal I$.
The $\ell_1$ norm of its curvelet coefficients is  
$
\norm{\mathcal C \{ \boldsymbol v \} }_{\ell_1}.
$
In order to get the sparse texture $\boldsymbol v$ in the spatial domain, 
the $\ell_1$ norm is adopted.
In conclusion, the norms $ \{ \norm{\mathcal C \{ \boldsymbol v \} }_{\ell_1} + \norm{\boldsymbol v}_{\ell_1} \} $ 
are considered to extract the fingerprint patterns.
Then, the bounded variation space with the discrete TV-norm, i.e. 
$ J(\boldsymbol u) = \norm{\nabla_\text{\tiny d} \boldsymbol u}_{\ell_1} $
(cf. \cite{AujolChambolle2005} for the definition of the discrete gradient operator $\nabla_\text{\tiny d}$), 
is well-known to measure the roughness of a piecewise constant image $\boldsymbol u$ \cite{RudinOsherFatemi1992}.
Finally, the residual $\boldsymbol \epsilon$ is measured by the supremum norm of its curvelet coefficients
, i.e. 
$$
\norm{ \mathcal C \{ \boldsymbol \epsilon \} }_{\ell_\infty}
= \sup_{i,l,\boldsymbol k \in \mathcal I} \abs{ \mathcal C_{i,l} \{\boldsymbol \epsilon\} [\boldsymbol k] }.
$$

Thus, the constraint of the minimisation is defined via
the supremum norm of the curvelet coefficients of the residual, i.e.
$\norm{ \mathcal C \{ \boldsymbol f - \boldsymbol u - \boldsymbol v \} }_{\ell_\infty}$, 
being less than a threshold $\delta$.
In summary, the variational solution we advocate for 
separating a fingerprint into texture, cartoon and noise in the Euclidean space $X$ 
whose dimension is given by the size of the lattice $\Omega$, i.e. $X = \mathbb R^{\abs{\Omega}}$, is defined as 
\begin{equation} \label{eq:OriMinimization0}
(\boldsymbol{\bar u}, \boldsymbol{\bar v}) ~=~ \argmin_{(\boldsymbol{u,v}) \in X^2} 
 \bigg\{ \norm{ \nabla_\text{\tiny d} \boldsymbol u }_{\ell_1} + \mu_1 \norm{ \mathcal C \{ \boldsymbol v \} }_{\ell_1} 
 + \mu_2 \norm{ \boldsymbol v }_{\ell_1} 
 \text{ s.t. } 
 \underbrace{ \sup_{i,l,\boldsymbol k \in \mathcal I} \abs{ \mathcal C_{i,l} \{\boldsymbol f - \boldsymbol u - \boldsymbol v \} [\boldsymbol k] } }_
 { =~ \norm{ \mathcal C \{ \boldsymbol f - \boldsymbol u - \boldsymbol v \} }_{\ell_\infty} }
 \leq \delta
 \bigg\} .
\end{equation}
Note that the form of (\ref{eq:OriMinimization0}) is analogous to the statistical multiresolution estimator in 
\cite{FrickMarnitzMunk2012} 
where the nonlinear transformation is the absolute value of the curvelet coefficients, i.e. $\Lambda(\cdot) = \abs{\mathcal C\{\cdot\}}$, 
the length of subsets $\abs{\mathcal S} = 1$ and the weight-function $\omega^{\mathcal S} = 1$.
The main difference is that our model has two variables $(\boldsymbol u \,, \boldsymbol v)$.
With the residual $\boldsymbol{\epsilon} = \boldsymbol{f} - \boldsymbol{u} - \boldsymbol{v} $, 
the constrained minimisation (\ref{eq:OriMinimization0}) is rewritten as
\begin{equation} \label{eq:OriMinimization}
 (\boldsymbol{\bar u}, \boldsymbol{\bar v}, \boldsymbol{\bar \epsilon}) ~=~ 
 \argmin_{(\boldsymbol{u}, \boldsymbol{v}, \boldsymbol{\epsilon}) \in X^3} \left\{ 
 \norm{ \nabla_\text{\tiny d} \boldsymbol{u} }_{\ell_1} + \mu_1 \norm{ \mathcal C \{ \boldsymbol{v} \} }_{\ell_1} + \mu_2 \norm{ \boldsymbol{v} }_{\ell_1} 
 \text{ s.t. } \norm{\mathcal C \{ \boldsymbol \epsilon \}}_{\ell_\infty} \leq \delta \,, 
 \boldsymbol{f} = \boldsymbol{u} + \boldsymbol{v} + \boldsymbol{\epsilon} 
 \right\} .
\end{equation}
Given $\delta > 0$, denote $G^* \big( \frac{\boldsymbol{\epsilon}}{\delta} \big)$ as the indicator function on the feasible convex set
$S(\delta)$ of (\ref{eq:OriMinimization}), i.e.
$$
S(\delta) ~=~ \bigg\{ \boldsymbol{\epsilon} \in X ~\mid~
\norm{\mathcal C \{ \boldsymbol \epsilon \}}_{\ell_\infty} \leq \delta \bigg\}
~~\text{and}~~
G^* \left( \frac{\boldsymbol{\epsilon}}{\delta} \right) ~=~ 
\begin{cases} 
 0, & \boldsymbol{\epsilon} \in S(\delta)  \\ 
 +\infty, & \boldsymbol{\epsilon} \in X \backslash S(\delta). 
\end{cases}
$$
By changing the inequality constraint into the indicator function $ G^* \left( \frac{\boldsymbol{\epsilon}}{\delta} \right) $, 
(\ref{eq:OriMinimization}) is rewritten as a convex minimisation of four convex functions and one equality constraint:
\begin{equation} \label{eq:min_constraint}
(\boldsymbol{\bar u}, \boldsymbol{\bar v}, \boldsymbol{\bar \epsilon}) ~=~ 
\argmin_{(\boldsymbol{u}, \boldsymbol{v}, \boldsymbol{\epsilon}) \in X^3} \left \{ 
\norm{\nabla_\text{\tiny d} \boldsymbol{u}}_{\ell_1} + \mu_1 \norm{ \mathcal C \{ \boldsymbol{v} \} }_{\ell_1} + 
\mu_2 \norm{\boldsymbol{v}}_{\ell_1} + G^* \left( \frac{\boldsymbol{\epsilon}}{\delta} \right)  
\text{ s.t. }  \boldsymbol{f} = \boldsymbol{u} + \boldsymbol{v} + \boldsymbol{\epsilon} 
\right\} .
\end{equation}
The original image $\boldsymbol{f}$ is therefore decomposed into 
the piecewise constant image $\boldsymbol{u}$, the texture $\boldsymbol{v}$
and the small scale objects modeling as noise $\boldsymbol{\epsilon}$
by minimizing the objective function (\ref{eq:min_constraint}).

\subsubsection{Augmented Lagrangian Method to Reformulate the Constrained Minimisation Problem in Equation (\ref{eq:min_constraint})}
\label{sssec:reformulating}

There are different kinds of norms in (\ref{eq:min_constraint}). 
In order to simplify the calculation, we introduce new variables
$$
\begin{cases}
 \boldsymbol{p} &=~ \nabla_\text{\tiny d} \boldsymbol{u} ~=~ \big[ \boldsymbol{p}_{1}\,, \boldsymbol{p}_{2} \big]^T
 \\
 \boldsymbol{w} &=~ \big[ \boldsymbol{w}_{i,l} \big] _{( i, l ) \in \mathcal I} 
 ~=~ \mathcal C \{ \boldsymbol{v} \} .
\end{cases}
$$

Then, (\ref{eq:min_constraint}) becomes a constrained minimisation and we apply the ALM. 
Given space $Y = X \times X$, the augmented Lagrangian 
function of (\ref{eq:min_constraint}) with the three Lagrange multipliers 
$(\boldsymbol{\lambda}_{\boldsymbol{1}} \,, \boldsymbol{\lambda}_{\boldsymbol{2}} \,, \boldsymbol{\lambda}_{\boldsymbol{3}})$ 
is defined as
\begin{equation} \label{eq:ALM2}
  \left( \boldsymbol{u}^* \,, \boldsymbol{v}^* \,, \boldsymbol{\epsilon}^* \,, \boldsymbol{w}^* 
  \,, \boldsymbol{p}^* \right) 
  ~=~ \argmin_{\boldsymbol{u} , \boldsymbol{v} , \boldsymbol{\epsilon} , \boldsymbol{w} , \boldsymbol{p} \in X^3 \times \mathbb R^{\abs{\mathcal I}} \times Y} 
  \mathcal L (\boldsymbol{u} \,, \boldsymbol{v} \,, \boldsymbol{\epsilon} \,, \boldsymbol{w} \,, \boldsymbol{p}; 
  \boldsymbol{\lambda}_{\boldsymbol{1}} \,, \boldsymbol{\lambda}_{\boldsymbol{2}} \,, \boldsymbol{\lambda}_{\boldsymbol{3}}) \,,
\end{equation}
where
\begin{align*}
 &\mathcal L (\boldsymbol{u}, \boldsymbol{v}, \boldsymbol{\epsilon}, \boldsymbol{w}, \boldsymbol{p}; 
             \boldsymbol{\lambda}_{\boldsymbol{1}}, \boldsymbol{\lambda}_{\boldsymbol{2}}, \boldsymbol{\lambda}_{\boldsymbol{3}}) 
 ~=~ \norm{ \boldsymbol{p} }_{\ell_1} + \mu_1 \norm{ \boldsymbol{w} }_{\ell_1} 
 + \mu_2 \norm{ \boldsymbol{v} }_{\ell_1} + G^* \left( \frac{\boldsymbol{\epsilon}}{\delta} \right) 
 \\
 &~~~ 
 + \frac{\beta_1}{2} \norm{ \boldsymbol{p} - \nabla_\text{\tiny d} \boldsymbol{u} 
 + \frac{\boldsymbol{\lambda}_{\boldsymbol{1}}}{\beta_1} }_{\ell_2}^2 
 + \frac{\beta_2}{2} \norm{ \boldsymbol{w} - \mathcal C \{ \boldsymbol{v} \} + 
    \frac{\boldsymbol{\lambda}_{\boldsymbol{2}}}{\beta_2} }_{\ell_2}^2 + 
    \frac{\beta_3}{2} 
    \norm{ \boldsymbol{f} - \boldsymbol{u} - \boldsymbol{v} - \boldsymbol{\epsilon} + \frac{\boldsymbol{\lambda}_{\boldsymbol{3}}}{\beta_3}}_{\ell_2}^2 .
\end{align*}
The minimizer of (\ref{eq:ALM2}) is numerically computed through iterations $n = 1 \,, 2 \,, \ldots$
\begin{equation} \label{eq:ALM3}
  \left( \boldsymbol{u}^{(n)} \,, \boldsymbol{v}^{(n)} \,, \boldsymbol{\epsilon}^{(n)} \,, \boldsymbol{w}^{(n)} \,, \boldsymbol{p}^{(n)} \right) 
  ~=~ \argmin_{\boldsymbol{u} , \boldsymbol{v} , \boldsymbol{\epsilon} , \boldsymbol{w} , \boldsymbol{p} \in X^3 \times \mathbb R^{\abs{\mathcal I}} \times Y} 
  \mathcal L (\boldsymbol{u} \,, \boldsymbol{v} \,, \boldsymbol{\epsilon} \,, \boldsymbol{w} \,, \boldsymbol{p}; 
  \boldsymbol{\lambda}_{\boldsymbol{1}}^{(n-1)} \,, \boldsymbol{\lambda}_{\boldsymbol{2}}^{(n-1)} \,, \boldsymbol{\lambda}_{\boldsymbol{3}}^{(n-1)})
\end{equation}
and the Lagrange multipliers are updated after every step $n$ with a rate $\gamma$ and the 
initial values $\boldsymbol \lambda_1^{(0)} = \boldsymbol \lambda_2^{(0)} = \boldsymbol \lambda_3^{(0)} = 0$:
$$
\begin{cases}
  \boldsymbol{\lambda}_{\boldsymbol{1}}^{(n)} ~=~ \boldsymbol{\lambda}_{\boldsymbol{1}}^{(n-1)} ~+~ 
  \gamma \beta_{1} ( \boldsymbol{p}^{(n)} - \nabla_\text{\tiny d} \boldsymbol{u}^{(n)} )
  \\
  \boldsymbol{\lambda}_{\boldsymbol{2}}^{(n)} ~=~ \boldsymbol{\lambda}_{\boldsymbol{2}}^{(n-1)} ~+~ 
  \gamma \beta_{2} ( \boldsymbol{w}^{(n)} - \mathcal C \{ \boldsymbol{v}^{(n)} \} )
  \\
  \boldsymbol{\lambda}_{\boldsymbol{3}}^{(n)} ~=~ \boldsymbol{\lambda}_{\boldsymbol{3}}^{(n-1)} ~+~ 
  \gamma \beta_{3} ( \boldsymbol{f} - \boldsymbol{u}^{(n)} - \boldsymbol{v}^{(n)} - \boldsymbol{\epsilon}^{(n)} )
\end{cases}.
$$
As the number of iterations $n$ goes to infinity, 
we obtain the true solution of (\ref{eq:ALM3}). 
However, to reduce the computational time in practice, we stop after a small number of iterations.
Hence, we gain an approximated solution (cf. Algorithm 1).

\begin{algorithm} \label{alg:ALMG3PD}
\caption{Augmented Lagrangian method (ALM) for the approximated solution of (\ref{eq:ALM2})}
\begin{algorithmic}
 \STATE Initialisation:
 $ 
 \boldsymbol{u}^{(0)} = \boldsymbol{v}^{(0)} = \boldsymbol{\epsilon}^{(0)} = \boldsymbol{p}^{(0)} =  
 \boldsymbol{w}^{(0)} = \boldsymbol{\lambda}_{\boldsymbol{1}}^{(0)} = \boldsymbol{\lambda}_{\boldsymbol{2}}^{(0)} = 
 \boldsymbol{\lambda}_{\boldsymbol{3}}^{(0)} = 0
 $
\FOR{$n=1$ to $N$}		
 \STATE 1. Compute the approximated solution
 $
 \big( \boldsymbol{u}^{(n)}, \boldsymbol{v}^{(n)}, \boldsymbol{\epsilon}^{(n)}, 
 \boldsymbol{w}^{(n)}, \boldsymbol{p}^{(n)} \big):
 $
 \begin{equation} \label{eq:ALM5}
  \big( \boldsymbol{u}^{(n)} , \boldsymbol{v}^{(n)} , \boldsymbol{\epsilon}^{(n)} , 
  \boldsymbol{w}^{(n)} , \boldsymbol{p}^{(n)} \big) ~=~ 
  \displaystyle \argmin_{\boldsymbol{u}, \boldsymbol{v}, \boldsymbol{\epsilon}, \boldsymbol{w}, \boldsymbol{p}} ~
  \mathcal L (\boldsymbol{u}, \boldsymbol{v}, \boldsymbol{\epsilon}, \boldsymbol{w}, \boldsymbol{p}; 
  \boldsymbol{\lambda}_{\boldsymbol{1}}^{(n-1)}, \boldsymbol{\lambda}_{\boldsymbol{2}}^{(n-1)}, \boldsymbol{\lambda}_{\boldsymbol{3}}^{(n-1)})
 \end{equation}

 \STATE 2. Update Lagrange multipliers
    $ \big( \boldsymbol{\lambda}_{\boldsymbol{1}}^{(n)}, \boldsymbol{\lambda}_{\boldsymbol{2}}^{(n)}, \boldsymbol{\lambda}_{\boldsymbol{3}}^{(n)} \big)$:
    \begin{equation} \label{eq:UpdateLambda}
    \begin{cases}	
     \boldsymbol{\lambda}_{\boldsymbol{1}}^{(n)} &=~ \boldsymbol{\lambda}_{\boldsymbol{1}}^{(n-1)} ~+~ 
     \gamma \beta_1 ( \boldsymbol{p}^{(n)} - \nabla_\text{\tiny d} \boldsymbol{u}^{(n)} ) 
     \\
     \boldsymbol{\lambda}_{\boldsymbol{2}}^{(n)} &=~ \boldsymbol{\lambda}_{\boldsymbol{2}}^{(n-1)} ~+~ 
     \gamma \beta_2 ( \boldsymbol{w}^{(n)} - \mathcal C \{\boldsymbol{v}^{(n)}\} )
     \\
     \boldsymbol{\lambda}_{\boldsymbol{3}}^{(n)} &=~ \boldsymbol{\lambda}_{\boldsymbol{3}}^{(n-1)} ~+~ 
     \gamma \beta_3 ( \boldsymbol{f} - \boldsymbol{u}^{(n)} - \boldsymbol{v}^{(n)} - \boldsymbol{\epsilon}^{(n)} )
    \end{cases}
    \end{equation}
\ENDFOR
\end{algorithmic}
\end{algorithm}


\begin{algorithm} \label{alg:ADMMG3PD}
\caption{Alternating direction method of multipliers (ADMM) for (\ref{eq:ALM5})}
\begin{algorithmic}
 \STATE  Fix Lagrange multipliers 
 $\boldsymbol{\lambda}_{\boldsymbol{1}} = \boldsymbol{\lambda}_{\boldsymbol{1}}^{(n-1)},
  \boldsymbol{\lambda}_{\boldsymbol{2}} = \boldsymbol{\lambda}_{\boldsymbol{2}}^{(n-1)}$
 and $\boldsymbol{\lambda}_{\boldsymbol{3}} = \boldsymbol{\lambda}_{\boldsymbol{3}}^{(n-1)}$,
 then alternatively solve the following sub-problems:   
 \begin{itemize}
  \item "$u$-problem":
         $ ~
         \boldsymbol{u}^{(n)} =~ \displaystyle \argmin_{\boldsymbol{u} \in X} ~ 
         \mathcal L (\boldsymbol{u}, \boldsymbol{v}^{(n-1)}, \boldsymbol{\epsilon}^{(n-1)}, \boldsymbol{p}^{(n-1)}, \boldsymbol{w}^{(n-1)};~ 
         \boldsymbol{\lambda}_{\boldsymbol{1}}, \boldsymbol{\lambda}_{\boldsymbol{2}}, \boldsymbol{\lambda}_{\boldsymbol{3}})             
         $
  \item "$v$-problem":
        $ ~
        \boldsymbol{v}^{(n)} =~ \displaystyle \argmin_{\boldsymbol{v} \in X} ~ 
        \mathcal L (\boldsymbol{u}^{(n)}, \boldsymbol{v}, \boldsymbol{\epsilon}^{(n-1)}, \boldsymbol{p}^{(n-1)}, \boldsymbol{w}^{(n-1)};~ 
        \boldsymbol{\lambda}_{\boldsymbol{1}}, \boldsymbol{\lambda}_{\boldsymbol{2}}, \boldsymbol{\lambda}_{\boldsymbol{3}})             
        $
  \item "$\epsilon$-problem":
        $ ~
        \boldsymbol{\epsilon}^{(n)} =~ \displaystyle \argmin_{\boldsymbol{\epsilon} \in X} ~ 
        \mathcal L (\boldsymbol{u}^{(n)}, \boldsymbol{v}^{(n)}, \boldsymbol{\epsilon}, \boldsymbol{p}^{(n-1)}, \boldsymbol{w}^{(n-1)};~ 
        \boldsymbol{\lambda}_{\boldsymbol{1}}, \boldsymbol{\lambda}_{\boldsymbol{2}}, \boldsymbol{\lambda}_{\boldsymbol{3}})             
        $
  \item "$p$-problem":
        $ ~
        \boldsymbol{p}^{(n)} =~ \displaystyle \argmin_{\boldsymbol{p} \in Y} ~ 
        \mathcal L (\boldsymbol{u}^{(n)}, \boldsymbol{v}^{(n)}, \boldsymbol{\epsilon}^{(n)}, \boldsymbol{p}, \boldsymbol{w}^{(n-1)};~ 
        \boldsymbol{\lambda}_{\boldsymbol{1}}, \boldsymbol{\lambda}_{\boldsymbol{2}}, \boldsymbol{\lambda}_{\boldsymbol{3}})             
        $
  \item "$w$-problem":
        $ ~
        \boldsymbol{w}^{(n)} =~ \displaystyle \argmin_{\boldsymbol{w} \in \mathbb R^{\abs{\mathcal I}}} ~ 
        \mathcal L (\boldsymbol{u}^{(n)}, \boldsymbol{v}^{(n)}, \boldsymbol{\epsilon}^{(n)}, \boldsymbol{p}^{(n)}, \boldsymbol{w};~ 
        \boldsymbol{\lambda}_{\boldsymbol{1}}, \boldsymbol{\lambda}_{\boldsymbol{2}}, \boldsymbol{\lambda}_{\boldsymbol{3}})             
        $         
  \end{itemize}  
\end{algorithmic}
\end{algorithm}

In the following part, we describe the algorithm to solve the minimisation problem (\ref{eq:ALM5}) by the alternating direction method of multipliers (ADMM).

\subsubsection{Alternating direction method of multipliers and numerical implementation}
\label{sssec:adm}

Similarly to \cite{TaiHahnChung2011}, \cite{LysakerLundervoldTai}, \cite{WuTai2010}, \cite{ZhuTaiChan2013}, \cite{ChanMarquinaMulet2000}, \cite{ZhuChan2012},
this section describes the procedure how to solve the minimisation (\ref{eq:ALM5}) and the method to discretize the solution. 

The solution of (\ref{eq:ALM5}) is determined by alternatively minimizing the objective function with respect to $\boldsymbol{u}$ 
while fixing $\boldsymbol{v}, \boldsymbol{\epsilon}, \boldsymbol{p}, \boldsymbol{w} $, and vice versa. 
Thus, we need to solve five subproblems denoted as "$ \boldsymbol{w} $-subproblem", 
"$ \boldsymbol{p} $-subproblem", "$ \boldsymbol{v} $-subproblem", 
"$ \boldsymbol{\epsilon} $-subproblem", "$ \boldsymbol{u} $-subproblem" as in Algorithm 2. 
The iterative scheme is as follows

{\bfseries ``$\boldsymbol{p}$-subproblem'':} Fix $\boldsymbol{u}, \boldsymbol{v}, \boldsymbol{\epsilon}, \boldsymbol{w}$ and
\begin{equation} \label{eq:psubproblem}
  \min_{\boldsymbol{p} \in Y} \left \{ \norm{ \boldsymbol{p} }_{\ell_1}  +  \frac{\beta_1}{2} \norm{ \boldsymbol{p} - \nabla_\text{\tiny d} \boldsymbol{u} 
  + \frac{\boldsymbol{\lambda}_{\boldsymbol{1}}}{\beta_1} }^2_{\ell_2} \right \}  
\end{equation}

Let $\nabla_\text{\tiny d} = [\partial_1^+,\partial_2^+]$ be the forward gradient operator \cite{AujolChambolle2005}.
The anisotropic version of (\ref{eq:psubproblem}) is solved by 
\begin{align} \label{eq:p} 
 \boldsymbol{\tilde p}_{1} ~=~ 
 \Shrink \left ( \partial^+_1 \boldsymbol{u} - \frac{\boldsymbol{\lambda}_{\boldsymbol{1},1}}{\beta_1}, 
                 \frac{1}{\beta_1} \right )
 ~~~~\text{and}~~~~
 \boldsymbol{\tilde p}_{2} ~=~ \Shrink \left ( \partial^+_2 \boldsymbol{u} 
 - \frac{\boldsymbol{\lambda}_{\boldsymbol{1},2}}{\beta_1}, \frac{1}{\beta_1} \right ),
\end{align}
where the shrinkage operator is defined as
\begin{equation*}
 \Shrink \big( x \,, \alpha \big) ~:=~ \frac{x}{\abs{x}} \cdot \max \big( \abs{x} - \alpha \,, 0 \big).
\end{equation*}

{\bfseries ``$\boldsymbol{w}$-subproblem'':} Fix $\boldsymbol{u}, \boldsymbol{v}, \boldsymbol{\epsilon}, \boldsymbol{p}$ and
\begin{equation} \label{eq:wsubproblem}
  \min_{ \boldsymbol{w} \in \mathbb R^{\abs{\mathcal I}} } \left \{ \mu_1 \norm{ \boldsymbol{w} }_{\ell_1} + \frac{\beta_2}{2} 
  \norm{ \boldsymbol{w} - \mathcal C \{ \boldsymbol{v} \} + \frac{\boldsymbol{\lambda}_{\boldsymbol{2}}}{\beta_2} }^2 _{\ell_2} \right \}  
\end{equation}

The solution of (\ref{eq:wsubproblem}) at the scale $i$ and the orientation $l$ is
\begin{equation} \label{eq:w}
 \boldsymbol{\tilde w}_{i,l} ~=~ \Shrink \left ( \mathcal C \{ \boldsymbol{v} \} - 
 \frac{\boldsymbol{\lambda}_{\boldsymbol{2},i,l}}{\beta_2}, \frac{\mu_1}{\beta_2}\right ),
 \qquad i, l \in \mathcal I.
\end{equation}

{\bfseries ``$\boldsymbol{v}$-problem'':} Fix $\boldsymbol{u}, \boldsymbol{\epsilon}, \boldsymbol{p}, \boldsymbol{w}$ and
\begin{equation} \label{eq:vsubproblem}
 \min_{ \boldsymbol{v} \in X } \left \{ \mu_2 \norm{ \boldsymbol{v} }_{\ell_1} + 
 \frac{\beta_2}{2} \norm{ \boldsymbol{w} - \mathcal C \{ \boldsymbol{v} \} + \frac{\boldsymbol{\lambda}_{\boldsymbol{2}}}{\beta_2}  
 }_{\ell_2}^2 + \frac{\beta_3}{2} 
 \norm{ \boldsymbol{f} - \boldsymbol{u} - \boldsymbol{v} - \boldsymbol{\epsilon} + 
 \frac{\boldsymbol{\lambda}_{\boldsymbol{3}}}{\beta_3} }^2_{\ell_2} \right \} 
\end{equation}

This (\ref{eq:vsubproblem}) is solved by 
\begin{equation} \label{eq:v}
 \boldsymbol{\tilde v} ~=~ \Shrink \left ( \boldsymbol{A}, \frac{\mu_2}{\beta_2 + \beta_3} \right ), \quad
\end{equation}
with
\begin{equation}
  \boldsymbol{A} ~=~ 
  \frac{ \mathcal C^* \big\{ \beta_2 \boldsymbol{w} + \boldsymbol{\lambda}_{\boldsymbol{2}} \big\} + 
  \beta_3 (\boldsymbol{f} - \boldsymbol{u} - \boldsymbol{\epsilon} + 
  \frac{\boldsymbol{\lambda}_{\boldsymbol{3}}}{\beta_3}) }{ \beta_2 + \beta_3 }.
  \label{eq:A_v}
\end{equation}

{\bfseries ``$\boldsymbol{\epsilon}$-problem'':} Fix $\boldsymbol{u}, \boldsymbol{v}, \boldsymbol{p}, \boldsymbol{w}$ and
\begin{equation}  \label{eq:epsilonsubproblem}
 \displaystyle \min_{\boldsymbol{\epsilon} \in X} \left\{ G^*\left( \frac{\boldsymbol{\epsilon}}{\delta} \right) + 
 \frac{\beta_3}{2} \norm{ \boldsymbol{\epsilon} - \left( \boldsymbol{f} - \boldsymbol{u} - \boldsymbol{v} + 
 \frac{\boldsymbol{\lambda}_{\boldsymbol{3}}}{\beta_3} \right) }^2_{\ell_2} \right\}
\end{equation}

This (\ref{eq:epsilonsubproblem}) is solved by (the proof is similar to \cite{AujolChambolle2005})
\begin{equation}     \label{eq:epsilon}
 \boldsymbol{\tilde \epsilon} ~=~ 
 \left( \boldsymbol{f} - \boldsymbol{u} - \boldsymbol{v} + \frac{\boldsymbol{\lambda}_{\boldsymbol{3}}}{\beta_3} \right) - 
 \CST \left( \boldsymbol{f} - \boldsymbol{u} - \boldsymbol{v} + \frac{\boldsymbol{\lambda}_{\boldsymbol{3}}}{\beta_3}, \delta \right),
\end{equation}
with the curvelet soft-thresholding:
$\CST ( x \,, \alpha ) ~:=~ \mathcal C^* \Big\{ \Shrink(\mathcal C\{x\} \,, \alpha) \Big\}.$

{\bfseries ``$\boldsymbol{u}$-problem'':} Fix $\boldsymbol{v}, \boldsymbol{p}, \boldsymbol{\epsilon}, \boldsymbol{w}$ and
\begin{equation}  \label{eq:usubproblem}
  \min_{ \boldsymbol{u} \in X } \left \{ \frac{\beta_1}{2} \norm{ \boldsymbol{p} - \nabla_\text{\tiny d} \boldsymbol{u} + 
         \frac{\boldsymbol{\lambda}_{\boldsymbol{1}}}{\beta_1} }^2_{\ell_2} + 
         \frac{\beta_3}{2} \norm{ \boldsymbol{f} - \boldsymbol{u} - \boldsymbol{v} - \boldsymbol{\epsilon} + 
         \frac{\boldsymbol{\lambda}_{\boldsymbol{3}}}{\beta_3} }^2_{\ell_2} \right \}
\end{equation}
Given the discrete finite frequency coordinates $\boldsymbol \omega = [\omega_1 \,, \omega_2] \in [-\pi \,, \pi]^2$
and let 
$F \big(e^{j \boldsymbol \omega}\big) \,, V \big(e^{j \boldsymbol \omega}\big)$, $E\big(e^{j \boldsymbol \omega}\big) \,,
\Lambda_3\big(e^{j \boldsymbol \omega}\big) \,, P_1\big(e^{j \boldsymbol \omega}\big) \,, P_2\big(e^{j \boldsymbol \omega}\big)$ 
and $\Lambda_1\big(e^{j \boldsymbol \omega}\big)$
be the discrete Fourier transform of 
$f[\boldsymbol k] \,, v[\boldsymbol k] \,, \epsilon[\boldsymbol k] \,, \lambda_3[\boldsymbol k] \,, p_1[\boldsymbol k] \,, p_2[\boldsymbol k]$
and $\lambda_1[\boldsymbol k]$, respectively.
This (\ref{eq:usubproblem}) is solved by 
\begin{equation*}
 \boldsymbol{\tilde u} ~=~
 \RE \Big[ \mathcal F^{-1} \Big\{ 
 \frac{ D\left( e^{j \boldsymbol \omega} \right) }{ \beta_3 + 4 \beta_1 \big[ \sin^2(\frac{\omega_1}{2}) + \sin^2(\frac{\omega_2}{2}) \big] }
 \Big\} \Big], 
\end{equation*}
with
\begin{align*} \label{eq:D}
D\big( e^{j \boldsymbol \omega} \big) 
 &~=~ \beta_3 \Big[ F\big( e^{j \boldsymbol \omega} \big)  
 - V \big( e^{j \boldsymbol \omega} \big) - E \big( e^{j \boldsymbol \omega} \big) 
 + \frac{ \Lambda_3 \big( e^{j \boldsymbol \omega} \big) }{\beta_3} \Big]             \notag
 \\
 &- \beta_1 \Big[  
 \big( 1 - e^{- j \omega_1} \big) 
 \Big( P_1 \big( e^{j \boldsymbol \omega} \big)
 + \frac{ \Lambda_{1,1} \big( e^{j \boldsymbol \omega} \big) }{\beta_1} \Big) 
 + \big( 1 - e^{- j \omega_2} \big) 
 \Big( P_2 \big( e^{j \boldsymbol \omega} \big)
 + \frac{ \Lambda_{1,2} \big( e^{j \boldsymbol \omega} \big) }{\beta_1} \Big)      
 \Big].
\end{align*}
The updated Lagrange multiplier $\boldsymbol \lambda_{\boldsymbol 1}^{(n)}$, $\boldsymbol \lambda_{\boldsymbol 2}^{(n)}$ and $\boldsymbol \lambda_{\boldsymbol 3}^{(n)}$
in (\ref{eq:UpdateLambda}) are
$$
\begin{cases}
 \boldsymbol{\lambda}_{\boldsymbol{1}, 1}^{(n)} ~=~ 
 \boldsymbol{\lambda}_{\boldsymbol{1},1}^{(n-1)} ~+~ 
 \gamma \beta_1 \left( \boldsymbol{\tilde p}_{1}^{(n)} - \partial_1^+ \boldsymbol{\tilde u}^{(n)} \right),
 \\
 \boldsymbol{\lambda}_{\boldsymbol{1},2}^{(n)} ~=~ \boldsymbol{\lambda}_{\boldsymbol{1},2}^{(n-1)} ~+~ 
 \gamma \beta_1 \left( \boldsymbol{\tilde p}_{2}^{(n)} - \partial_2^+ \boldsymbol{\tilde u}^{(n)}  \right),
 \\
 \boldsymbol{\lambda}_{\boldsymbol{2},i,l}^{(n)} ~=~ \boldsymbol{\lambda}_{\boldsymbol{2},i,l}^{(n-1)} ~+~ \gamma \beta_2 
 \left( \boldsymbol{\tilde w}_{i,l}^{(n)} - \mathcal C_{i,l} \{ \boldsymbol{\tilde v}^{(n)} \} \right),
 \\
 \boldsymbol{\lambda}_{\boldsymbol{3}}^{(n)} ~=~ \boldsymbol{\lambda}_{\boldsymbol{3}}^{(n-1)} ~+~ \gamma \beta_3 
 \left( \boldsymbol{f} - \boldsymbol{\tilde u}^{(n)} - \boldsymbol{\tilde v}^{(n)} - \boldsymbol{\tilde \epsilon}^{(n)} \right) .
\end{cases}
$$

For a given $ \gamma $ in Algorithm 1, the solution of (\ref{eq:min_constraint}) is obtained by applying 
alternatively the above formulas in the subproblems. This is a convex 
program with the alternating minimisation procedure. However, the choice of parameters $(\mu_1,\mu_2,\delta)$ and 
$(\beta_1, \beta_2, \beta_3)$ affects on the solution. 
Since the texture information is an essential feature for the segmentation process, 
the parameter $\mu_1$ and $\mu_2$ are important, especially $ \mu_2 $ 
controls the sparsity of the fingerprint texture. 
In this context, $\mu_2$ is adaptively designed to cancel $(\beta_2 + \beta_3)$ in the shrinkage operator 
(\ref{eq:v}), it depends only on the maximum of $A$ and the constant $C$, as follows
\begin{equation}
  \mu_2 ~=~ C ( \beta_2+\beta_3 ) \cdot \max_{\boldsymbol k \in \Omega}(A[\boldsymbol k]),
  \label{eq:mu2}
\end{equation}
where $ A[\boldsymbol k] $ is defined in (\ref{eq:A_v}). 
Since the fingerprint images are captured by various kinds of sensors, 
their properties and qualities differ. 
Therefore, $C$ is obtained empirically from training sets for each type of sensor.

The parameter $ \delta $ is used to remove the small scale objects (noise). 
In order to reduce these kinds of noise in $\boldsymbol v$
such that $\boldsymbol v$ contains mainly fingerprint pattern as good as possible
(cf. Figure \ref{fig:overviewG3DP}), we simply approximate these noise as Gaussian.
See \cite{DelosreyesSchoenlieb2013} for a combination of multiple noise models, 
e.g. considering Gaussian, Poisson and impulse noise, simultaneously.
According to the extreme value behavior of the curvelet coefficients (cf. \cite{HaltmeierMunk2014}),
the threshold $\delta$ is chosen with the quantile $\alpha = 0.7$ from the asymptotic distribution as 
\begin{equation}
 \delta = \sigma \sqrt{2 \log \abs{\mathcal I}} + \sigma \frac{ 2z - \log \log \abs{\mathcal I} - \log \pi }{ 2 \sqrt{2\log\abs{\mathcal I}} }
 ~~\text{ and }~~
 z = -\log \log \big( \frac{1}{1-\alpha} \big),
 \label{eq:delta}
\end{equation}
where $\abs{\mathcal I}$ is total number of curvelet coefficients and $\sigma$ is commonly calculated from the first level 
of the Cohen-Daubechies-Feauveau 9/7 wavelet high-frequency diagonal coefficient ($HH1$) 
(cf. \cite{CohenDaubechiesFeauveau1992}):
$$
\sigma ~=~ \frac{\text{median} (~\abs{HH1}~)}{0.6745}.
$$
Note that this approximation depends on the normality assumption of the noise, which may not always be true in practice.
This can be adapted to different noise models as in general the threshold
can always be obtained via simulation.

\subsubsection{Smoothness and Sparsity of the Extracted Texture}				
\label{sssec:smoothsparse}

\begin{figure*}[t]
\begin{center}
 \subfigure[$\boldsymbol f$]{ \includegraphics[width=0.22\textwidth]{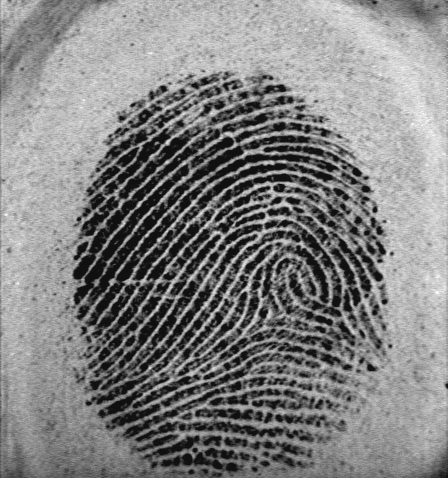} }		
 \subfigure[$F(e^{j\boldsymbol \omega})$]{ \includegraphics[width=0.22\textwidth]{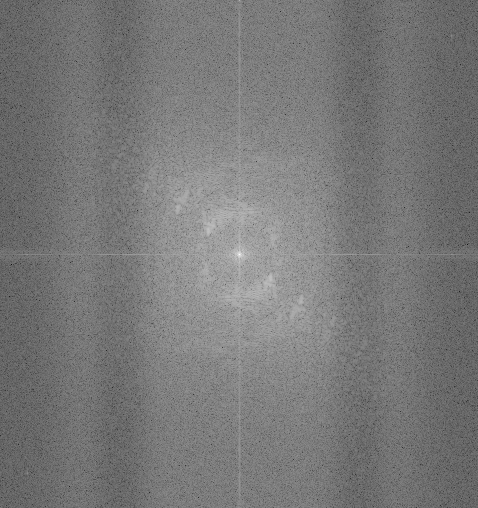} }
 \subfigure[$\boldsymbol u$]{ \includegraphics[width=0.22\textwidth]{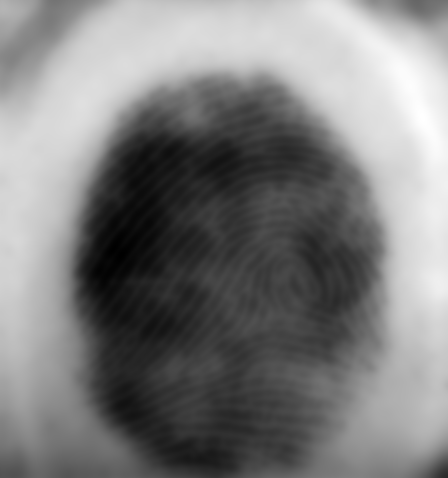} }
 \subfigure[$\boldsymbol v$]{ \includegraphics[width=0.22\textwidth]{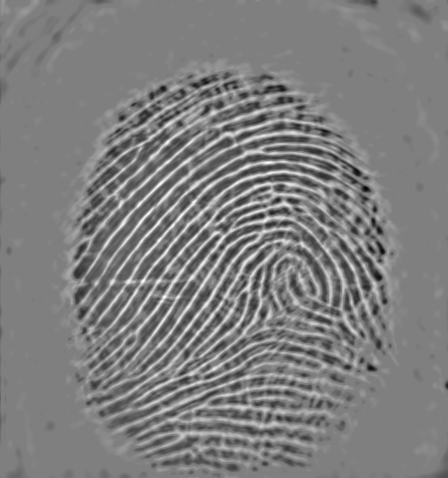} }

 \subfigure[$\boldsymbol \epsilon$]{ \includegraphics[width=0.22\textwidth]{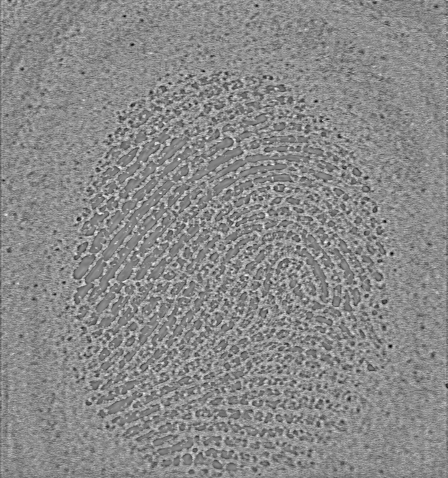} }		
 \subfigure[$U(e^{j\boldsymbol \omega})$]{ \includegraphics[width=0.22\textwidth]{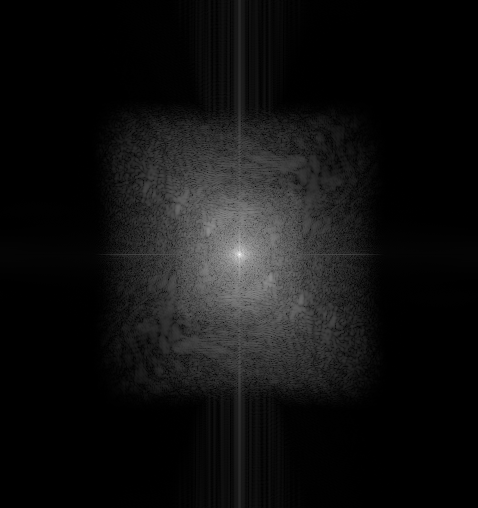} }	
 \subfigure[$V(e^{j\boldsymbol \omega})$]{ \includegraphics[width=0.22\textwidth]{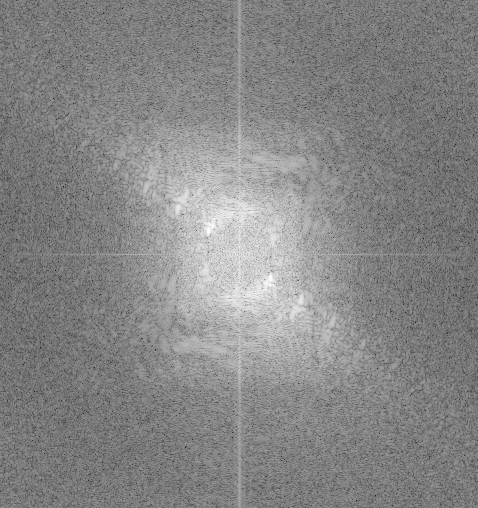} }		
 \subfigure[$E(e^{j\boldsymbol \omega})$]{ \includegraphics[width=0.22\textwidth]{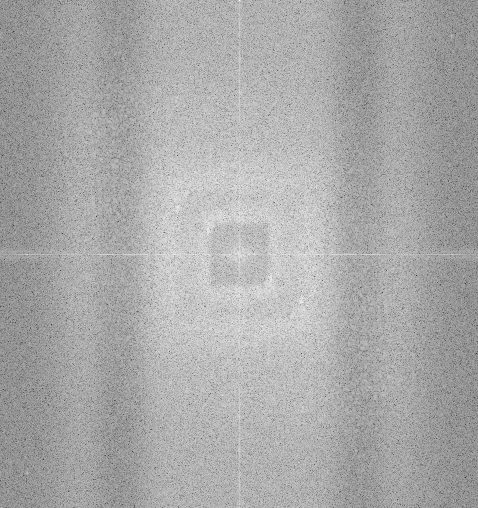} }		

 \caption{ \small A fingerprint image and its Fourier spectrum are shown in (a) and (b), respectively.
									Image (a) is decomposed by G3PD with $\mu_1 = 1$, iteration = 20, level = 5,
                  $\beta_1 = 0.06 \,, \beta_2 = \beta_3 = \gamma = 10^{-3}$
									into cartoon (c), texture (d) and noise (e) images.
									Their respective Fourier spectra are visualized in (f-h).
									We observe that the Fourier spectra of the component images 									
									resemble responses after lowpass, bandpass and highpass filtering.			
									Please note that especially (d) and (g) show that the fingerprint pattern is mostly concentrated in 
                  a specific range of frequencies \cite{ThaiHuckemannGottschlich2015}.
         }
 \label{fig_Imagedecompose}
\end{center}
\end{figure*}

The oscillation signal corresponding to fingerprint patterns is considered as
a sparse and smooth texture which is decomposed by the G3PD model of the original fingerprint image $\boldsymbol f$ into three parts 
satisfying the constraint $\boldsymbol f = \boldsymbol u + \boldsymbol v + \boldsymbol \epsilon$ (cf. Figure \ref{fig_Imagedecompose}), including:
the piecewise-constant image $\boldsymbol u$, the texture $\boldsymbol v$, and noise $\boldsymbol \epsilon$.

\begin{figure*}[ht]
\begin{center}
 \subfigure[]{ \includegraphics[width=0.16\textwidth,height=0.24\textwidth]{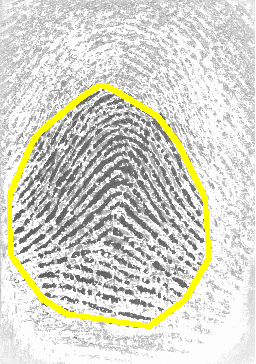} }
 \subfigure[$k=3$]{ \includegraphics[width=0.16\textwidth,height=0.24\textwidth]{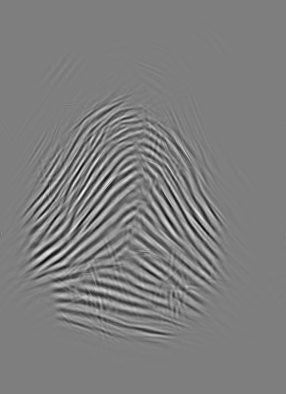} }
 \subfigure[$k=6$]{ \includegraphics[width=0.16\textwidth,height=0.24\textwidth]{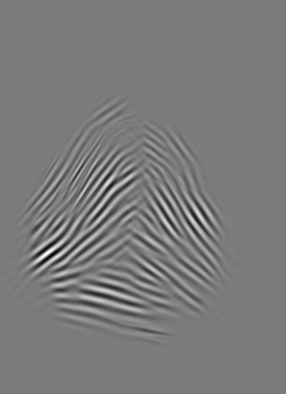} }
 \subfigure[$k=8$]{ \includegraphics[width=0.16\textwidth,height=0.24\textwidth]{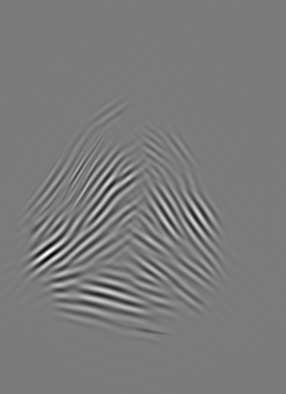} }
 \subfigure[$k=20$]{ \includegraphics[width=0.16\textwidth,height=0.24\textwidth]{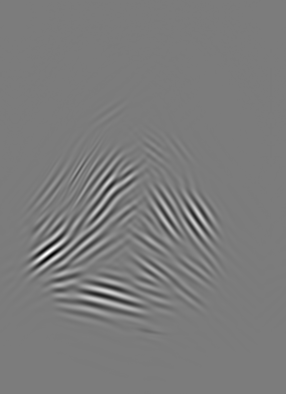} }
 
 \subfigure[]{ \includegraphics[width=0.16\textwidth,height=0.24\textwidth]{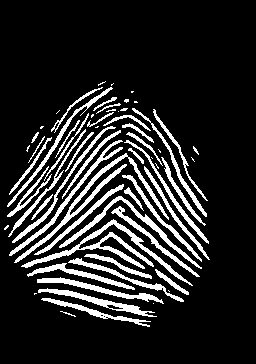} }
 \subfigure[]{ \includegraphics[width=0.16\textwidth,height=0.24\textwidth]{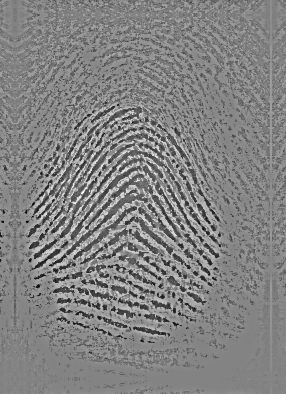} }
 \subfigure[]{ \includegraphics[width=0.16\textwidth,height=0.24\textwidth]{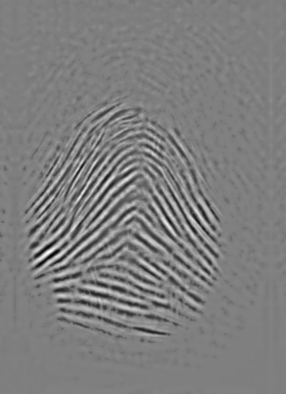} }
 \subfigure[]{ \includegraphics[width=0.16\textwidth,height=0.24\textwidth]{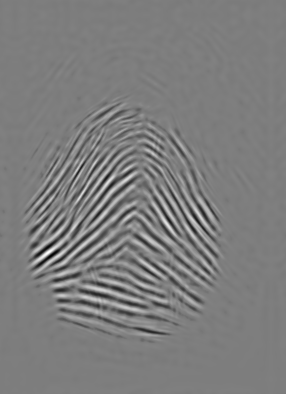} }
 \subfigure[]{ \includegraphics[width=0.16\textwidth,height=0.24\textwidth]{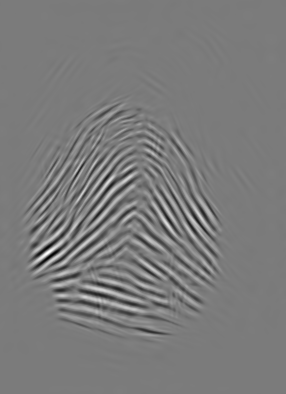} }
 \caption{ \small Image (a) depicts the original image where the yellow line indicates 
                  the boundary of the ROI estimated by the G3PD method after 20 iterations.
									The ROI is obtained using morphological operations on the binarized image~(f).
									Images (b-e) show $\boldsymbol v_{\text{smooth}}$ in (\ref{eq:RewriteTexture}), 
									the smoothing term of $\boldsymbol v$ and
									(g-j) visualize the corresponding $\boldsymbol v_{\text{update}}$ in (\ref{eq:RewriteTexture})
									after $k$ iterations.
         }
 \label{fig:DecompEffectOf2ShrinkageOptAndUpdateOfTexture}
\end{center}
\end{figure*}

In this section, we will analyze how the norms $\norm{\mathcal C \{\boldsymbol v\}}_{\ell_1}$ and $\norm{\boldsymbol v}_{\ell_1}$ in (\ref{eq:min_constraint})
affect on the smoothness and sparsity of the extracted texture
which is our main goal for the feature extraction.
In order to do that, a closed form of $\boldsymbol v$ is found by 
putting (\ref{eq:w}) and (\ref{eq:A_v}) into (\ref{eq:v}),
letting $\theta = \frac{\beta_2}{\beta_2 + \beta_3}$ 
and the thresholds $T_1 = \frac{\mu_1}{\beta_2}$ and $T_2 = \frac{\mu_2}{\beta_2 + \beta_3}$:
\begin{equation} \label{eq:RewriteTexture}
 \boldsymbol{\tilde v} ~=~ \Shrink \Bigg( \theta \underbrace{ \mathcal C^* \bigg\{ \Shrink \Big( \mathcal C\{\boldsymbol v\} 
              - \frac{\boldsymbol{\lambda_2}}{\beta_2} \,, T_1 \Big) 
              + \frac{\boldsymbol{\lambda_2}}{\beta_2} \bigg\} }_{ \displaystyle :=~ \boldsymbol{v}_{\text{smooth}} ~\approx~ \CST \big( \boldsymbol v \,, T_1 \big) }
              ~+~ (1 - \theta) \underbrace{ \bigg( \boldsymbol f - \boldsymbol u - \boldsymbol \epsilon + \frac{\boldsymbol{\lambda_3}}{\beta_3} \bigg) }_{ \displaystyle :=~ \boldsymbol v_{\text{update}} }
              \,, T_2\Bigg).
\end{equation}
We see that the estimated texture $\boldsymbol{\tilde v}$ contains two shrinkage operators: respectively, the inside and the outside correspond to 
the smoothness and sparseness terms resulting from $\norm{\mathcal C{\boldsymbol v}}_{\ell_1}$ and $\norm{\boldsymbol v}_{\ell_1}$ in (\ref{eq:min_constraint}). 
(cf. Figure \ref{fig:DecompEffectOf2ShrinkageOptAndUpdateOfTexture} 
for the effects of the smoothness and sparseness of $\boldsymbol v$ after different numbers of iterations).
These effects can be observed in the binarised texture (Figure \ref{fig:DecompEffectOf2ShrinkageOptAndUpdateOfTexture} (f)).
The parameter $\theta \in (0 \,, 1)$ in (\ref{eq:RewriteTexture}) serves as a regularisation parameter to balance 
between the smoothing term $\boldsymbol v_{\text{smooth}}$ and the updated term $\boldsymbol v_{\text{update}}$.

\begin{figure*}[ht]
\begin{center}
 \subfigure{ \includegraphics[width=0.20\textwidth]{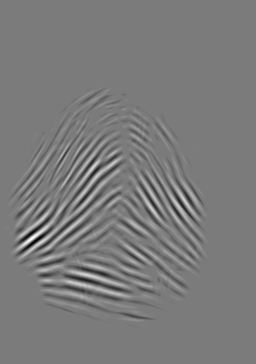} }
 \subfigure{ \includegraphics[width=0.20\textwidth]{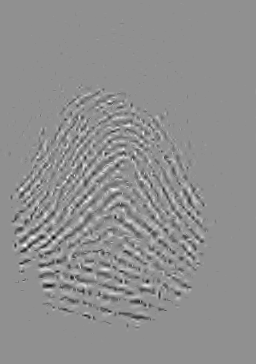} }
 \subfigure{ \includegraphics[width=0.20\textwidth]{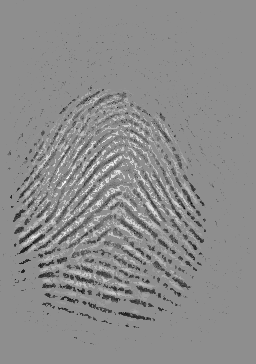} }
 \subfigure{ \includegraphics[width=0.20\textwidth]{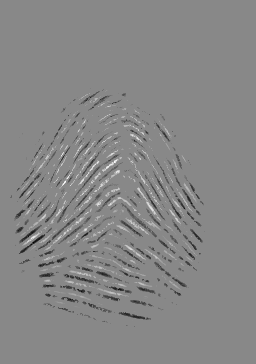} }
 
 \subfigure{ \includegraphics[width=0.20\textwidth]{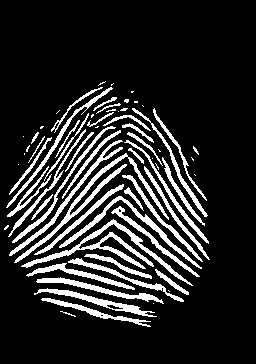} }
 \subfigure{ \includegraphics[width=0.20\textwidth]{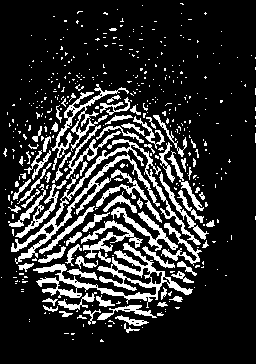} }
 \subfigure{ \includegraphics[width=0.20\textwidth]{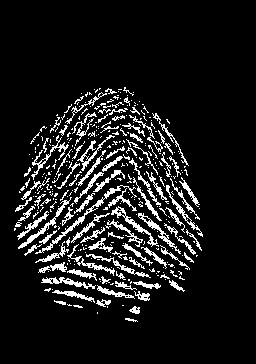} }
 \subfigure{ \includegraphics[width=0.20\textwidth]{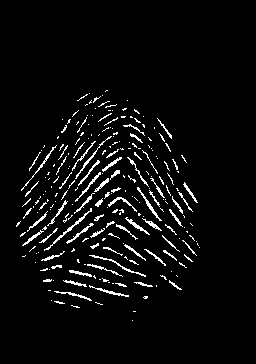} }

   \subfigure{ \includegraphics[width=0.20\textwidth]{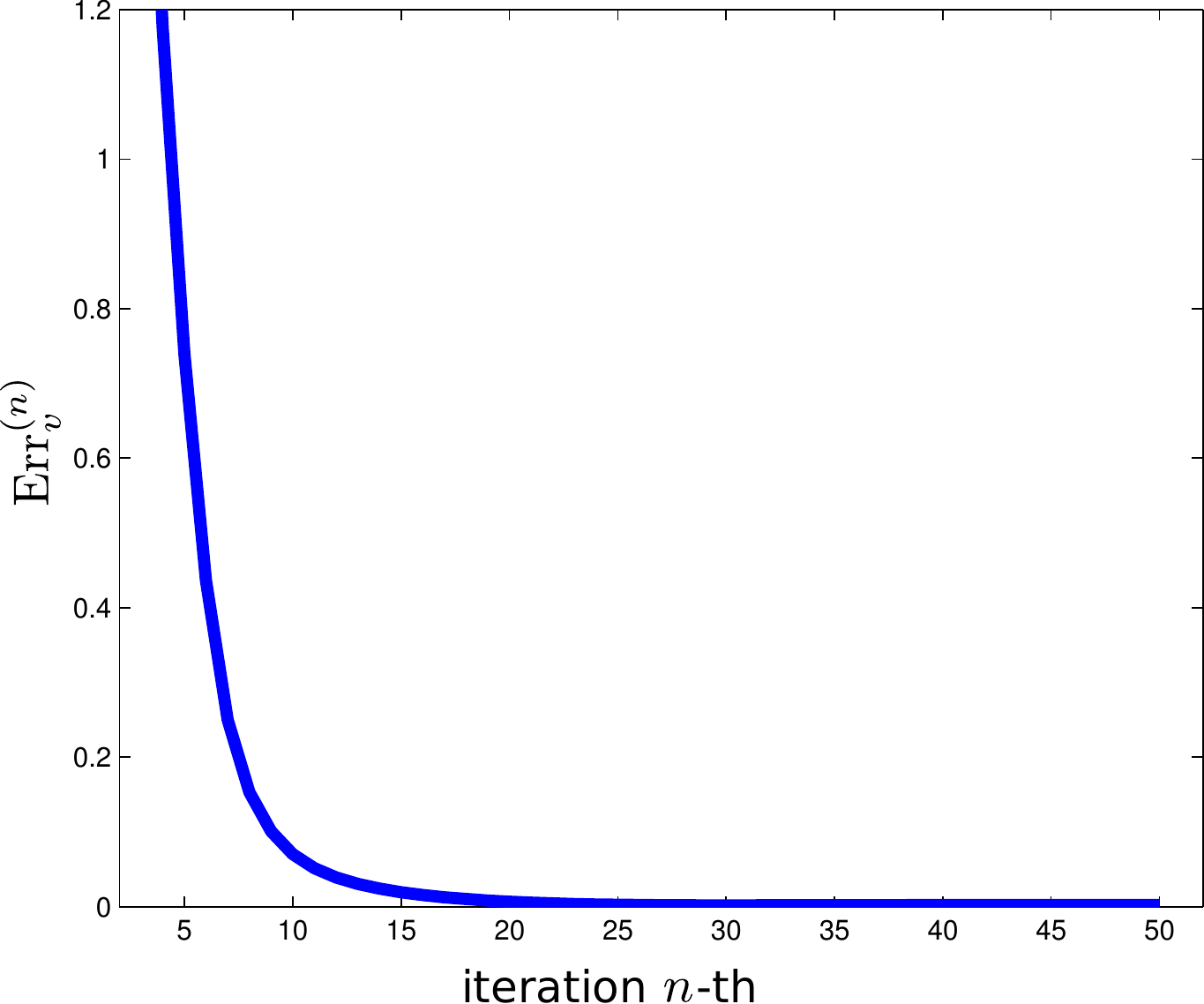} }
   \subfigure{ \includegraphics[width=0.20\textwidth]{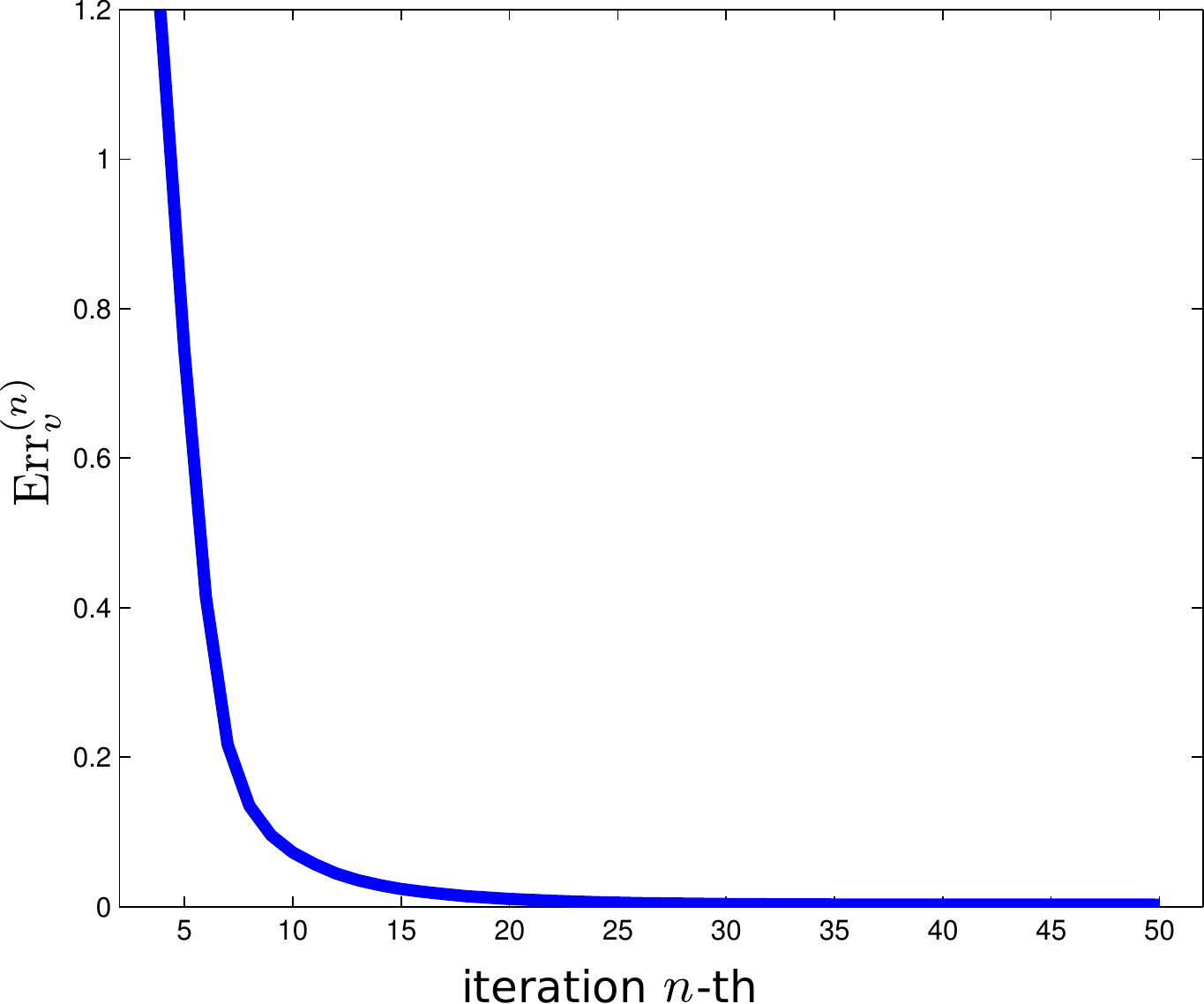} }
   \subfigure{ \includegraphics[width=0.20\textwidth]{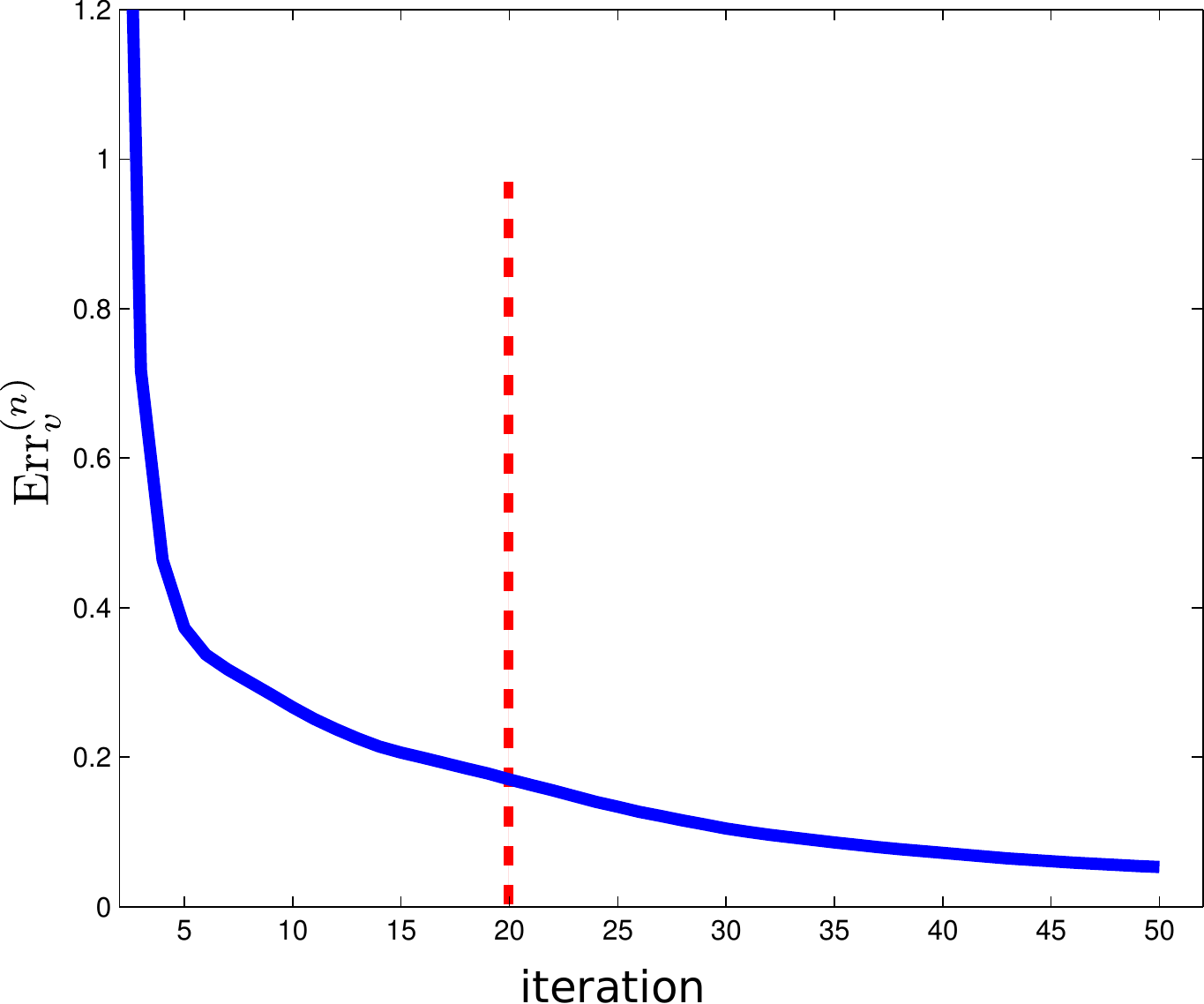} }
	 \subfigure{ \includegraphics[width=0.20\textwidth]{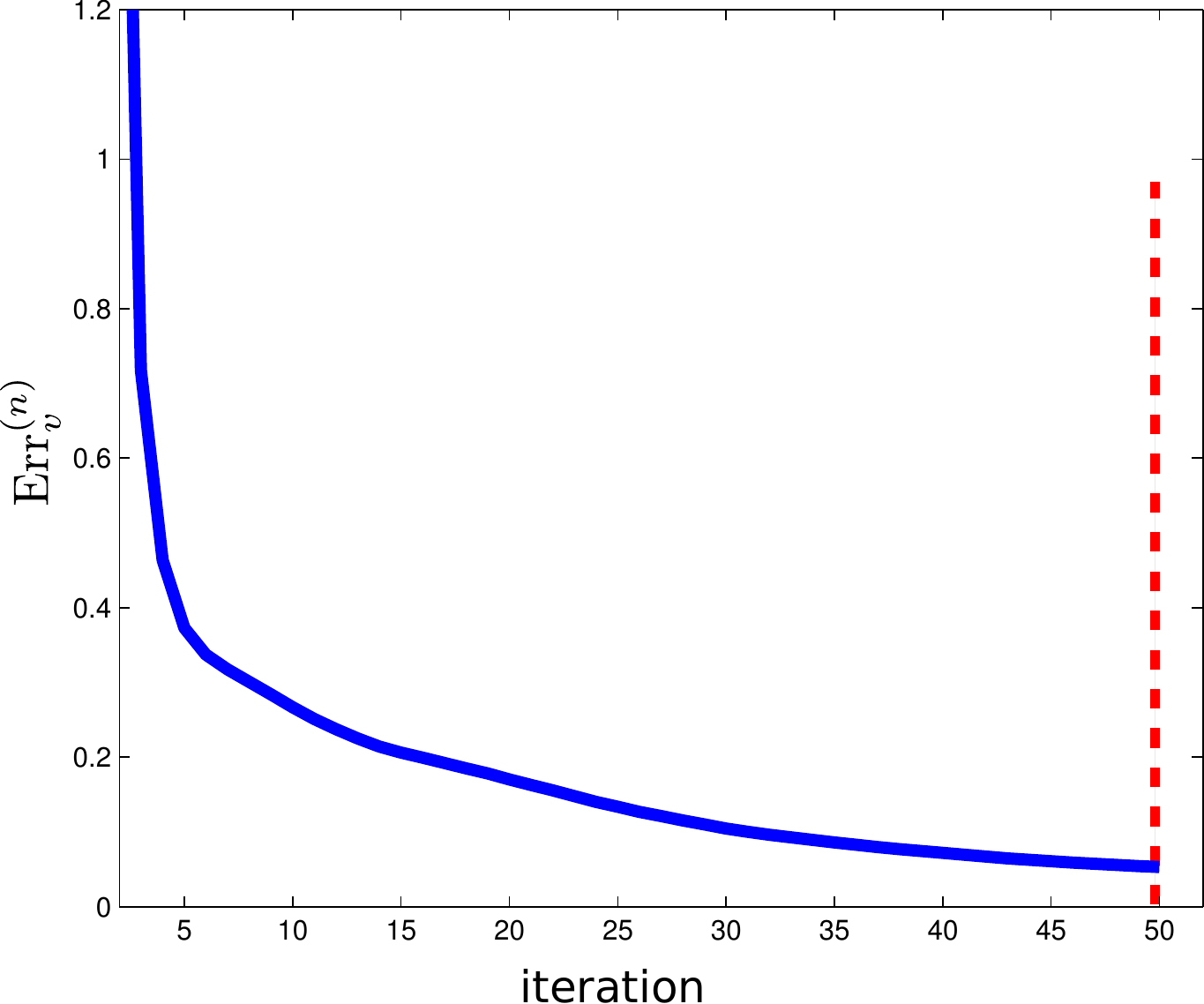} }
  
 \caption{ \small The following comparison illustrates the effect of the smoothness term for $\boldsymbol v$ in (\ref{eq:OriMinimization0}).
                  The first column depicts the $\ell_1$ norm of curvelet coefficients, i.e. $\norm{\mathcal C\{\boldsymbol v\}}_{\ell_1}$
									and the second column the $\ell_1$ norm of wavelet coefficients, i.e. $\norm{\mathcal W\{\boldsymbol v\}}_{\ell_1}$
									after 50 iterations.
									Columns 3 and 4 visualize $\boldsymbol v$ obtained without smoothness term 
									(no $\norm{\mathcal C\{\boldsymbol v\}}_{\ell_1}$ in (\ref{eq:OriMinimization0}))
									after 20 and 50 iterations, respectively.									
									The first row shows the texture images $\boldsymbol v$, 
									the second row their binarized versions
									and the third row their plots of convergence rates.									
									The comparison shows that the curvelet based smoothness term 
									leads to a better texture image than the wavelet based one
									and that convergence without smoothness term is slow and texture tends to be destroyed.
         }
 \label{fig:ComparisonProposedWaveNoCur20NoCurv50}
\end{center}
\end{figure*}

Figure \ref{fig:ComparisonProposedWaveNoCur20NoCurv50} compares 
the effect of the curvelet smoothness term 
$\norm{\mathcal C\{\boldsymbol v\}}_{\ell_1}$ in (\ref{eq:OriMinimization0}) with a wavelet smoothness term 
$\norm{\mathcal W\{\boldsymbol v\}}_{\ell_1}$ and without smoothness measurement.
The texture $\boldsymbol v$ estimated using $\norm{\mathcal W\{\boldsymbol v\}}_{\ell_1}$ in (\ref{eq:OriMinimization0})
is not as good regarding smoothness and sparseness as the texture obtained by the curvelet smoothness term.
In order to evaluate the convergence rate of the algorithm, 
we denote the relative error between successive iterations as
\begin{equation} \label{eq:RelativeErrV}
 \displaystyle \text{Err}^{(n)}_{\boldsymbol v} = \frac{ \norm{\boldsymbol v^{(n)} - \boldsymbol v^{(n-1)}}_{\ell_2} }{ \norm{\boldsymbol v^{(n-1)}}_{\ell_2} }.
\end{equation}
In Figure \ref{fig:ComparisonProposedWaveNoCur20NoCurv50}, one can see that 
without smoothness measurement, the convergence rate is slow (cf. 3rd row) 
and the algorithm tends to eliminate texture (cf. column 3 and 4). 
Note that for the smoothness measurement $\norm{\mathcal C\{\boldsymbol v\}}_{\ell_1}$, 
the proposed method achieves a stable estimated texture $\boldsymbol v$ after circa 20 iterations. 
Hence, the estimated $\boldsymbol v$ and its binarisation
are almost the same
after 20 or 50 iterations
(see Figure \ref{fig:DecompEffectOf2ShrinkageOptAndUpdateOfTexture} (f) and 
1st column of Figure \ref{fig:ComparisonProposedWaveNoCur20NoCurv50}).


\subsection{Morphological Operations}
\label{ssec:morphology}

Firstly, the smooth and sparse texture $\boldsymbol v$ is extracted by the combination of the $\ell_1$ norm of its curvelet coefficients
and its $\ell_1$ norms, simultaneously.
Secondly, post-processing as described in \cite{ThaiHuckemannGottschlich2015} is applied to obtain the ROI. 
More specifically, the morphological operations act only on the non-zero coefficients of the texture image $\boldsymbol v$.
In other words, this corresponds to a projection of the thresholding value to the parameter $\mu_2$ 
which has been designed to adapt to the intensity of each image by Eq. (\ref{eq:mu2}).

\section{Evaluation: Benchmark, Protocol and Experimental Results} 
\label{sec:evaluation}

\subsection{Benchmark and Evaluation Metric}

The publicly available fingerprint images of the FVC competitions from 2000, 2002 and 2004
are used as benchmark for evaluating segmentation performance.
Each competition consists of four databases: three databases are acquired from real fingers
and the fourth database of each competition is synthetically generated.

It has recently been shown that real and synthetic fingerprints can be discriminated with very high accuracy
using minutiae histograms (MHs) \cite{GottschlichHuckemann2014}.  
More specifically, by computing the MH for a minutiae template 
and then computing the earth mover's distance (EMD) \cite{GottschlichSchuhmacher2014} 
between the MH of the template and the mean MHs for a set of real and synthetic fingerprints.
Classification is simply performed by choosing the class with the smaller EMD. 

In total, there are 12 databases and each database contains 880 images (80 for training 
and 800 for testing).
The ground truth segmentation has been manually marked for these 10560 images 
as described in \cite{ThaiHuckemannGottschlich2015}.

Let $N_1$ and $N_2$ be the width and height of image $\boldsymbol f$ in pixels.
Let $M_f$ be number of pixels which are marked as foreground by human experts
and estimated as background by an algorithm (missed/misclassified foreground). 
Let $M_b$ be number of pixels which are marked as background by human experts
and estimated as foreground by an algorithm (missed/misclassified background). 
The average total error per image is defined as
\begin{equation} \label{eq:segmentationerror}
 Err = \frac{M_f + M_b}{N_1 \times N_2}.
\end{equation}

\subsection{Parameter Selection}

Parameters for all methods considered in the comparison 
are selected on the training set of 80 images for each database.
More specifically, those parameters are chosen which minimize 
the segmentation error defined in (\ref{eq:segmentationerror}) for the respective training set.
Choosing the parameters for each database is appropriate,
because the nine databases consisting of real fingerprints
have been acquired using nine different sensors
and the images of each database have sensor-specific properties.
The parameter selection for the FDB \cite{ThaiHuckemannGottschlich2015}, GFB \cite{ShenKotKoo2001}, HCR \cite{WuTulyakovGovindaraju2007},
MVC \cite{BazenGerez2001} and STFT \cite{ChikkerurCartwrightGovindaraju2007} methods
are discussed in \cite{ThaiHuckemannGottschlich2015}.

\begin{table}
\begin{center}
  \begin{tabular}{|c|l|}
    \hline
    Parameters		& Description
    \\
    \hline
    $N$			& the number of iterations in the Algorithm 1.
    \\
    \hline
    $\mu_1$		& the regularised parameter for $\ell_1$ norm of curvelet coefficients $\mathcal C\{\boldsymbol v\}$ 
    \\
                       & in Eq. (\ref{eq:OriMinimization}).  
    \\
    \hline
    $C$			& the adaptive constant in Eq. (\ref{eq:mu2}) for the regularised parameter $\mu_2$ 
    \\
                       & in $\ell_1$ norm of $\boldsymbol v$ in Eq. (\ref{eq:OriMinimization}).    
    \\
    \hline
    $\beta_1 \,, \beta_2 \,, \beta_3$	& the parameters in the augmented Lagrangian function (\ref{eq:ALM2}). 
    \\
    \hline
    $\gamma$   		& the rate of the updated Lagrange multipliers in Eq. (\ref{eq:ALM2}). 
    \\
    \hline
    $s$   		& the window size of the block in the postprocessing step in \cite[Eq. (8)]{ThaiHuckemannGottschlich2015}.
    \\
    \hline
    $t$   		& a constant for selecting the morphology threshold $T$ in \cite[Eq. (8)]{ThaiHuckemannGottschlich2015}.
    \\
    \hline
    $b$   		& the number of the neighbouring blocks in \cite[Eq. (8)]{ThaiHuckemannGottschlich2015}.
    \\
    \hline
    $p$   		& the mirror boundary condition to avoid the boundary effect.
    \\
    \hline
  \end{tabular}    
  \vspace*{8pt}
  \caption{Overview over all parameters for the global three-part decomposition (G3PD) method for fingerprint segmentation. 
           Values are reported in Table \ref{tabParameterChoice}.
          \label{tabParameters}
          } 
\end{center}
\end{table}

\begin{table}
\begin{center}
  \begin{tabular}{|c|c|c|c|c|}
    \hline
    FVC		& DB		& $C$	& $\beta_2$	
    \\
    \hline
    2000	& 1		& 0.045		& 0.0005		
    \\
	        & 2		& 0.045		& 0.0100		
    \\
		& 3		& 0.055		& 0.0010		
    \\
	        & 4		& 0.025		& 0.0010	
    \\
    \hline
    2002	& 1		& 0.020		& 0.0010		
    \\
	        & 2		& 0.035		& 0.0005		
    \\
		& 3		& 0.070		& 0.0010		
    \\
		& 4		& 0.020		& 0.0500		
    \\
    \hline
    2004	& 1		& 0.015		& 0.1000		
    \\
	        & 2		& 0.025		& 0.0010		
    \\
		& 3		& 0.035		& 0.0010		
    \\
	        & 4		& 0.035		& 0.0005		
    \\
    \hline
  \end{tabular} 
  \vspace*{8pt}
  \caption{Overview over the parameters learned on the training set.
         The other eight parameters are $\mu_1 = 1$, $\beta_1 = \beta_3 = \gamma = 10^{-3}$, $s = 9$, $t = 5$, $b = 6$ and $p = 15$
         for all databases.
         \label{tabParameterChoice}}
\end{center}
\end{table}

For the proposed G3PD method, the involved parameters are summarized in Table~\ref{tabParameters}
and the values of the learned parameters are reported in Table~\ref{tabParameterChoice}.
In a reasonable amount of time, a number of conceivable parameter combinations were tried on the training set. 

For different numbers of iterations, we have applied the following training scheme:
\begin{itemize}
 \item Firstly, $C$, an adaptive constant for $\mu_2$ in (\ref{eq:mu2}) to define a threshold for the sparseness of $\boldsymbol v$, 
       is trained while fixing the other parameters. 
 \item Secondly, with the obtained $C$, we train the other parameters one by one while fixing the rest.
\end{itemize}

The two parameters which have the biggest impact on the segmentation performance 
are the number of iterations $N$ and the constant $C$ in Eq. (\ref{eq:mu2}).
Therefore, these two parameters have been trained first.
In our experiments, the minimum error on the training set averaged over all 12 databases is obtained 
for $N=4$ iterations. 
In these practical applications of our proposed model, stopping before convergence leads to better segmentation results 
which are also influenced by the combination with the morphological operations. 
For further details and a discussion, see \cite{Thai2015PhD}.

Note that the solution of $(\boldsymbol u \,, \boldsymbol v \,, \boldsymbol \epsilon) $ depends severely on the choices of $(\mu_1 \,, \mu_2 \,, \delta)$, 
as well as the parameters of the
optimisation step $(\beta_1 \,, \beta_2 \,, \beta_3 \,, \gamma)$. 
To achieve a good decomposition in which cartoon, texture and noise 
are separated 
is difficult in practice, because there are no models of noise and texture. 
Fortunately, this paper 
focuses on the segmentation of fingerprint images for which the texture $\boldsymbol v$ is important. 
After the decomposition, there can still be pattern contents in the cartoon image $\boldsymbol u$ 
and the noise image $\boldsymbol \epsilon$ (see Figure \ref{fig:overviewG3DP}), 
but the important aspect is that the texture image $\boldsymbol v$ is adequate for segmentation.

The choice of aforementioned parameters balances the amount of pattern in the texture image
with the smoothness of the cartoon image. 
Selecting parameters which increase the smoothness of the cartoon image $\boldsymbol u$,
also tend to cause the halo effect in the texture image $\boldsymbol v$.
We observe that especially $\beta_1$ influences this trade-off:
if $\boldsymbol u$ contains only homogeneous regions (cf. Figure \ref{fig_Imagedecompose} (c)), 
it tends to generate the halo effect on the boundary of fingerprint pattern in $\boldsymbol v$ (cf. Figure \ref{fig_Imagedecompose} (d)).
Particularly, the halo effect results from the blurred homogeneous region $\boldsymbol u$.
In order to reduce this effect in $\boldsymbol v$, the parameters are chosen 
such that the algorithm assigns ``enough'' texture to $\boldsymbol v$. 
Hence, $\boldsymbol u$ and $\boldsymbol \epsilon$ can contain some partial textures, 
but this yields better a segmentation performance, cf. \cite{Thai2015PhD}.

\begin{figure*}[ht]
\begin{center}
 \subfigure[]{ \includegraphics[width=0.19\textwidth]{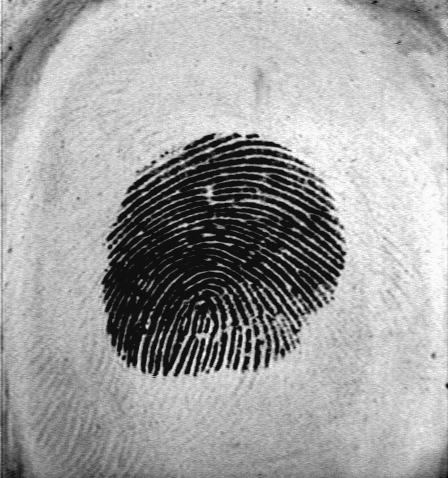} }
 \subfigure[G3PD: $\beta_1 = 0.005$]{ \includegraphics[width=0.19\textwidth]{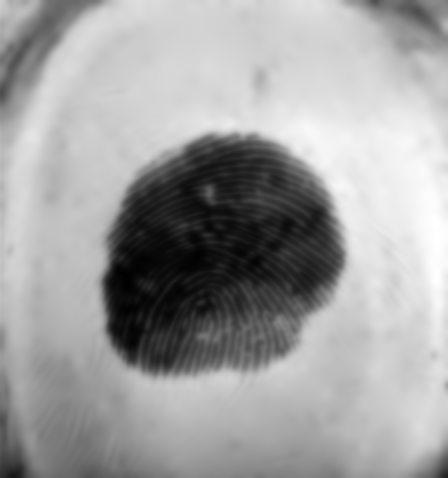} } 
 \subfigure[]{ \includegraphics[width=0.19\textwidth]{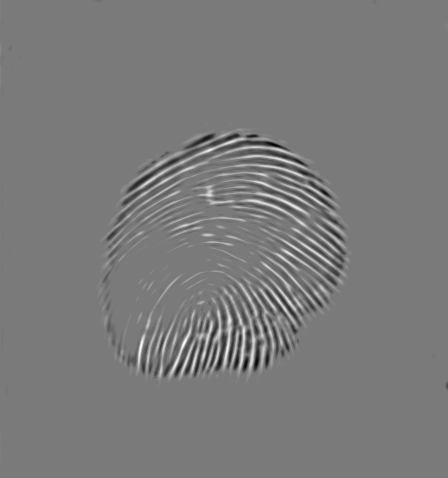} }
 \subfigure[]{ \includegraphics[width=0.19\textwidth]{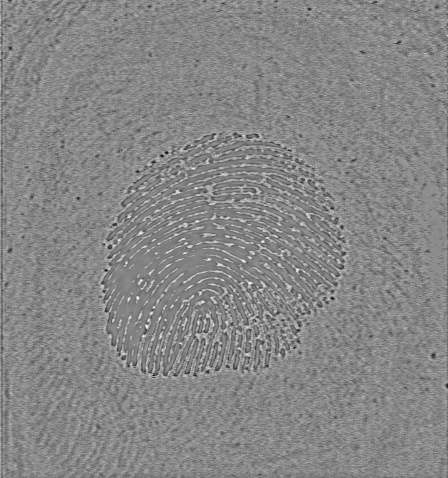} }
 
 \subfigure[$\lambda_{\text{TV}L_2} = 0.1$]{ \includegraphics[width=0.19\textwidth]{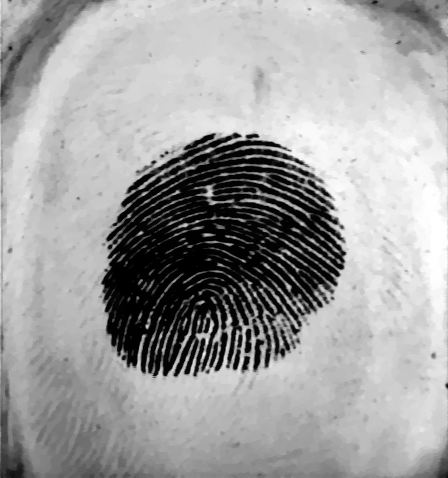} }
 \subfigure[]{ \includegraphics[width=0.19\textwidth]{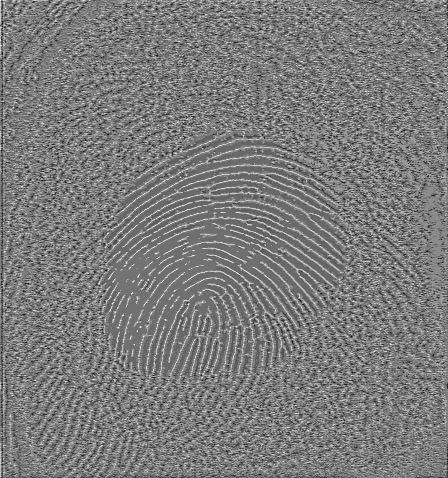} } 
 \subfigure[$\lambda_{\text{TV}L_1} = 1$]{ \includegraphics[width=0.19\textwidth]{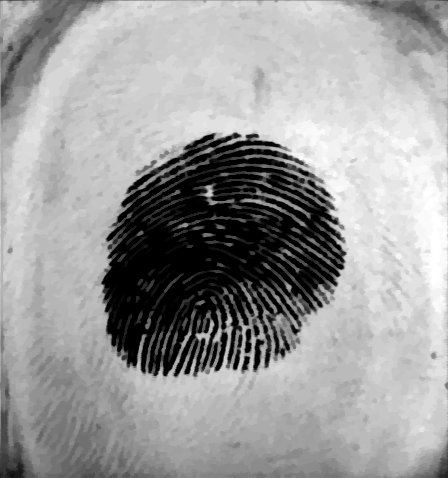} }
 \subfigure[]{ \includegraphics[width=0.19\textwidth]{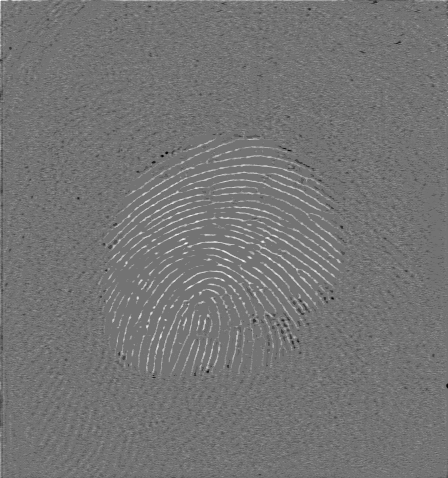} }
 
 \subfigure[$\lambda_{\text{TV}L_2} = 0.02$]
 { \includegraphics[width=0.19\textwidth]{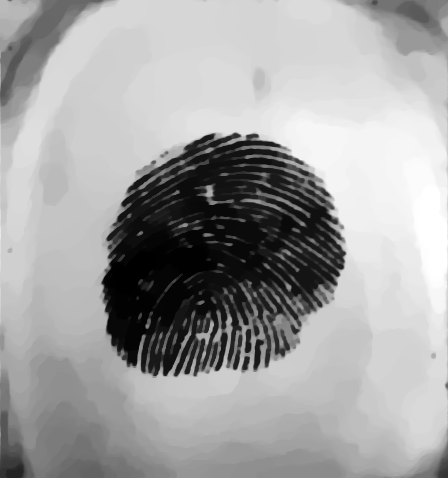} }
 \subfigure[]{ \includegraphics[width=0.19\textwidth]{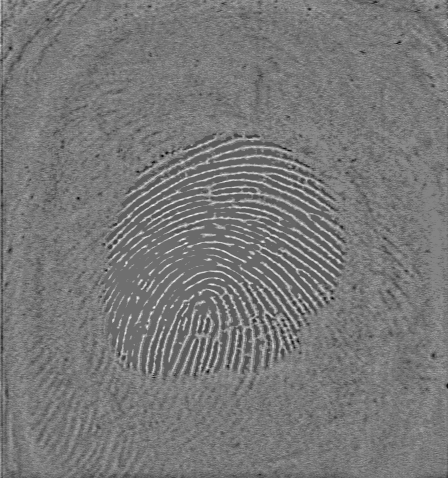} } 
 \subfigure[$\lambda_{\text{TV}L_1} = 0.6$]{ \includegraphics[width=0.19\textwidth]{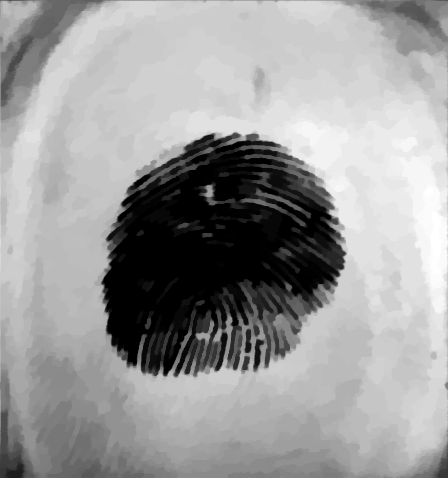} }
 \subfigure[]{ \includegraphics[width=0.19\textwidth]{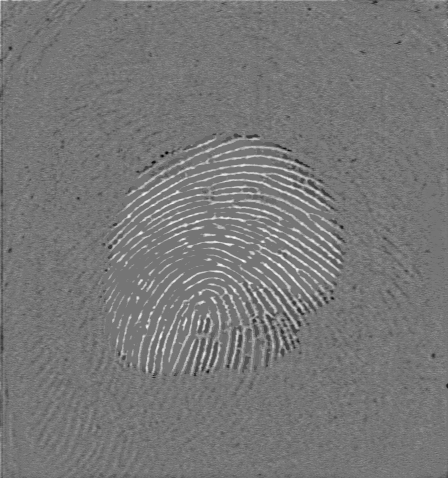} }
 
 \subfigure[$\lambda_{\text{TV}L_2} = 0.005$]
 { \includegraphics[width=0.19\textwidth]{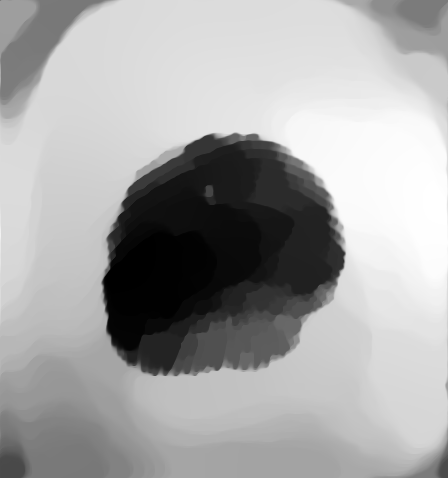} }
 \subfigure[]{ \includegraphics[width=0.19\textwidth]{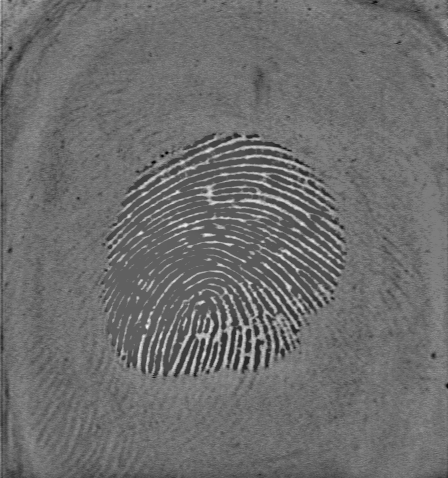} } 
 \subfigure[$\lambda_{\text{TV}L_1} = 0.1$]{ \includegraphics[width=0.19\textwidth]{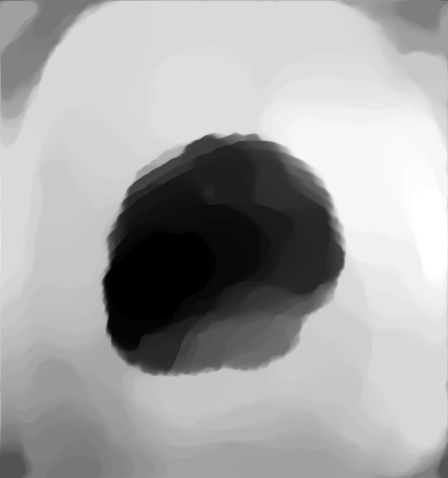} }
 \subfigure[]{ \includegraphics[width=0.19\textwidth]{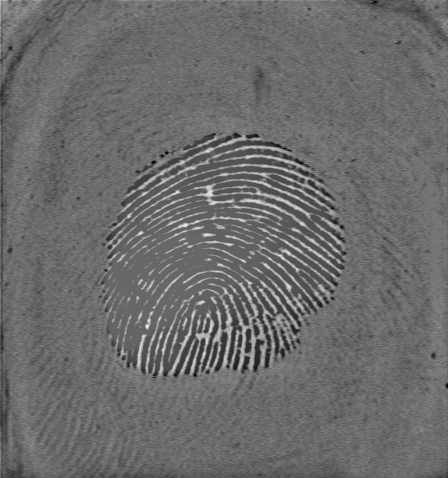} }

 \caption{ \small A comparison of G3PD with $\text{TV}-L_2$ and $\text{TV}-L_1$: 
                  First row: images f.l.t.r are the original image $\boldsymbol f$ 
                  and the three-part decomposition by G3PD with $N = 50 \,, \beta_1 = 0.005$
                  (see Table \ref{tabParameterChoice} for the other parameters): 
                  the cartoon image $\boldsymbol u$, texture image $\boldsymbol v$ and noise image $\boldsymbol \epsilon$.
                  The first and second column of rows two to four show 
                  images $\boldsymbol u$ and $\boldsymbol v$, respectively, 
									for $\text{TV}-L_2$ two-part decomposition with different values of $\lambda_{\text{TV}L_2}$.
                  The third and fourth column show the corresponding 
									images $\boldsymbol u$ and $\boldsymbol v$ for $TV-L_1$ two-part decomposition.
                  The number of iterations for $\text{TV}-L_2$ and $\text{TV}-L_1$ is $N = 350$.
                  Note that for no choice of $\lambda_{\text{TV}L_2}$ or $\lambda_{\text{TV}L_1}$, 
                  $\text{TV}-L_2$ or $\text{TV}-L_1$ produce a good feature image for segmentation of this noisy fingerprint,
                  while the G3PD model provides a useful texture image $\boldsymbol v$ for the segmentation procedure.            
         }     
 \label{fig:ComparisonWithTVL12}
\end{center}
\end{figure*}

Let us consider the comparison of the proposed model with the standard ROF $\text{TV}-L_2$ model \cite{RudinOsherFatemi1992} 
and the $\text{TV}-L_1$ model \cite{ChanEsedoglu2005} for feature decomposition (see Figure \ref{fig:ComparisonWithTVL12}).
For simplicity, let $\lambda_{\text{TV}L_2}$ and $\lambda_{\text{TV}L_1}$ be the regularisation parameters for
$\text{TV}-L_2$ and $\text{TV}-L_1$, respectively.
The ROF $\text{TV}-L_2$ model has been introduced by \cite{RudinOsherFatemi1992} 
for the purpose of image denoising. The ROF model has been designed to obtain a smooth cartoon image $\boldsymbol u$.
For fingerprint image segmentation we are interested in a texture image 
which is as useful as possible in terms of a feature for segmentation. 
However, the ROF model or the $\text{TV}-L_1$ model cannot produce a sparse and smooth texture image 
from a noisy fingerprint image $\boldsymbol f$
no matter how the corresponding parameter is selected. 
On the one hand, if the ROF model decomposes $\boldsymbol f$ into a very smooth cartoon image $\boldsymbol u$, 
than $\boldsymbol v$ contains both noise and texture. 
On the other hand, for a different choice of $\lambda_{\text{TV}L_2}$ or $\lambda_{\text{TV}L_1}$, 
$\boldsymbol v$ contains mostly noise
and $\boldsymbol u$ includes texture and large scale objects. 
In neither of the two situations, $\boldsymbol u$ or $\boldsymbol v$ is useful as a feature for fingerprint segmentation.
A comparison of the G3PD method with $\text{TV}-L_1$ and $\text{TV}-L_2$ two-part decomposition 
is shown in Figure \ref{fig:ComparisonWithTVL12}.
Zhang \textit{et al.} \cite{ZhangLaiKuo2013} have tried to solve this problem 
by proposing a locally adaptive two-part decomposition which also takes the orientation of the pattern into account.

\begin{figure*}[ht]
\begin{center}   
    \subfigure{ \includegraphics[width=0.16\textwidth]{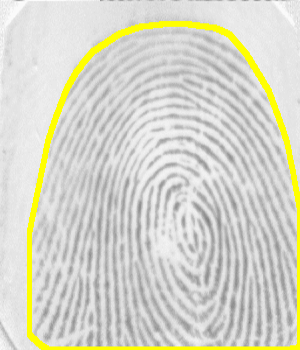} }
    \subfigure{ \includegraphics[width=0.16\textwidth]{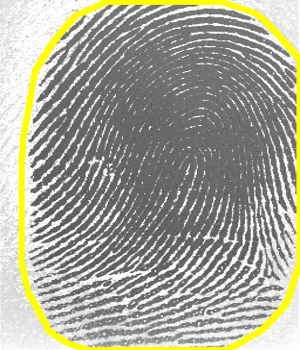} }
    \subfigure{ \includegraphics[width=0.16\textwidth]{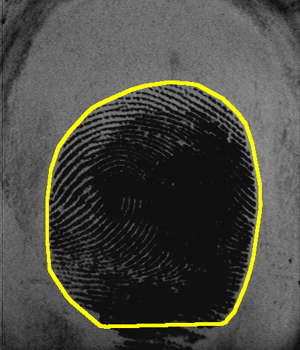} }
    \subfigure{ \includegraphics[width=0.16\textwidth]{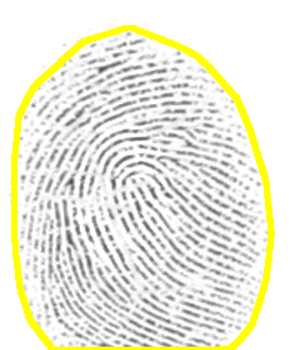} }
    \\
    \subfigure{ \includegraphics[width=0.16\textwidth]{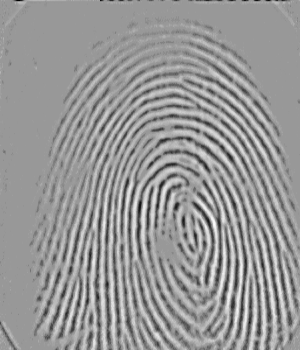} }
    \subfigure{ \includegraphics[width=0.16\textwidth]{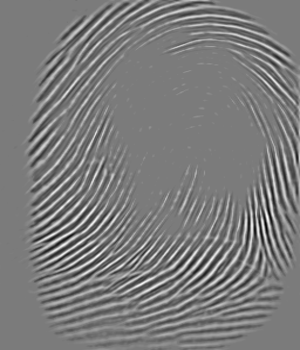} }
    \subfigure{ \includegraphics[width=0.16\textwidth]{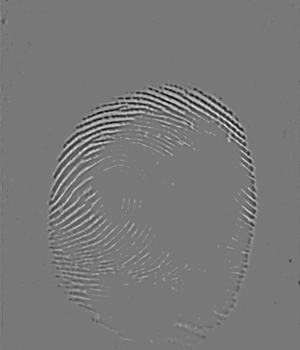} }
    \subfigure{ \includegraphics[width=0.16\textwidth]{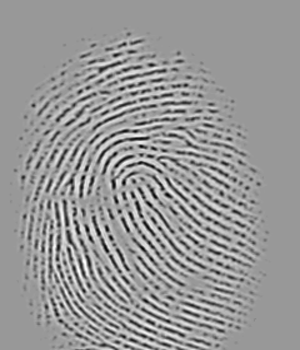} }    
    \\
    \subfigure{ \includegraphics[width=0.16\textwidth]{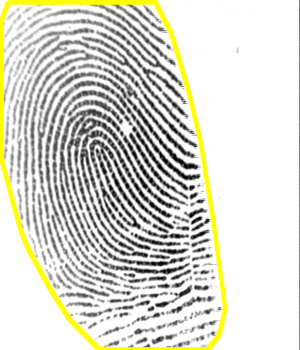} }
    \subfigure{ \includegraphics[width=0.16\textwidth]{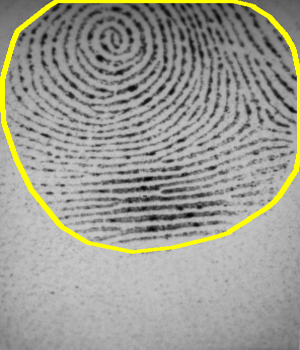} }
    \subfigure{ \includegraphics[width=0.16\textwidth]{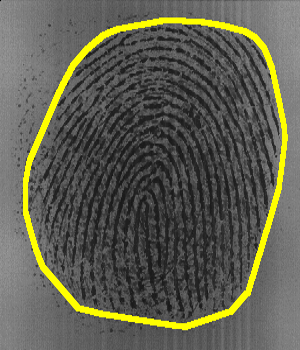} }
    \subfigure{ \includegraphics[width=0.16\textwidth]{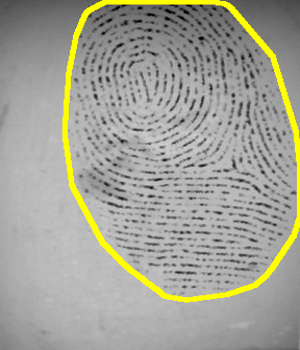} }
    \\
    \subfigure{ \includegraphics[width=0.16\textwidth]{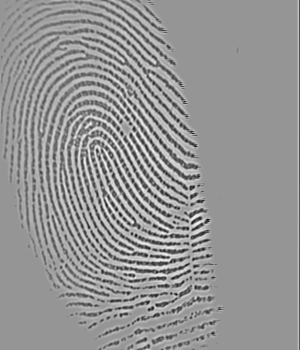} }
    \subfigure{ \includegraphics[width=0.16\textwidth]{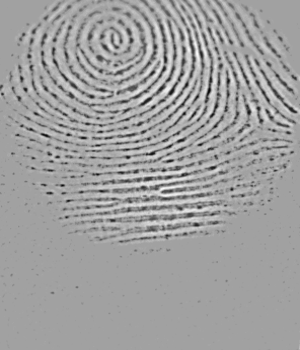} }
    \subfigure{ \includegraphics[width=0.16\textwidth]{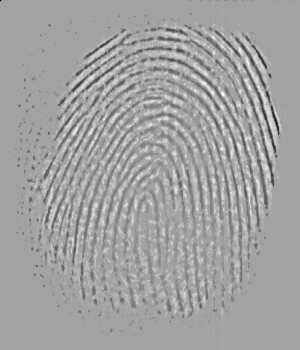} }
    \subfigure{ \includegraphics[width=0.16\textwidth]{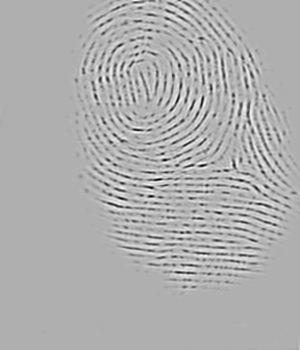} }    
    \\
    \subfigure{ \includegraphics[width=0.16\textwidth]{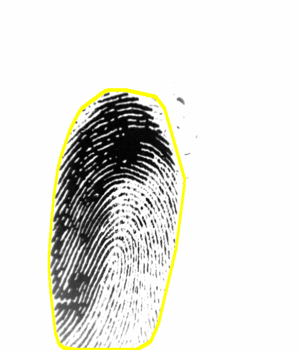} }
    \subfigure{ \includegraphics[width=0.16\textwidth]{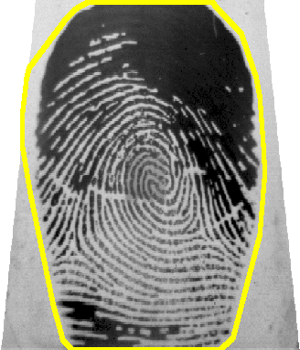} }
    \subfigure{ \includegraphics[width=0.16\textwidth]{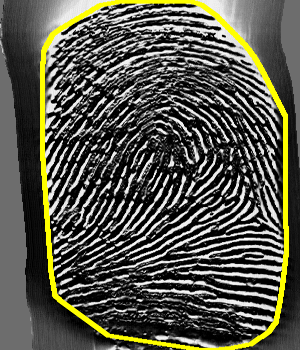} }
    \subfigure{ \includegraphics[width=0.16\textwidth]{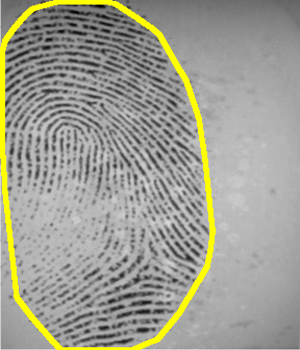} }
    \\
    \subfigure{ \includegraphics[width=0.16\textwidth]{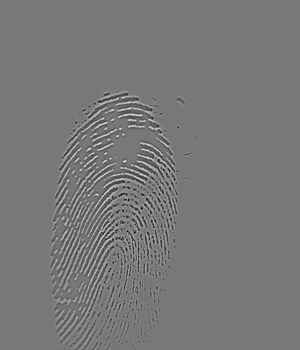} }
    \subfigure{ \includegraphics[width=0.16\textwidth]{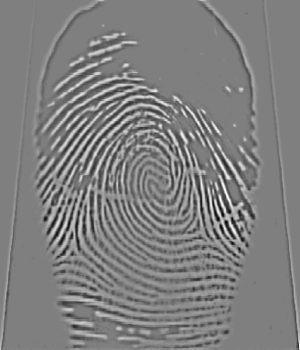} }
    \subfigure{ \includegraphics[width=0.16\textwidth]{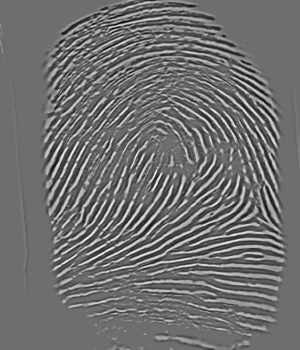} }
    \subfigure{ \includegraphics[width=0.16\textwidth]{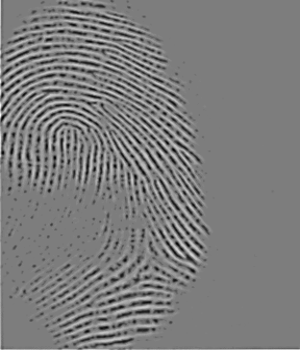} }
    
    \caption{Segmented fingerprint images and the corresponding texture images 
              by the variational method
              for FVC2000 (first and second row), FVC2002 (third and fourth row) 
 	      and FVC2004 (fifth and sixth row).
              Columns f.l.t.r correspond to DB1 to DB4.              
     \label{figSegmentationResultDHBBOverview}}
\end{center}
\end{figure*}

In summary, the proposed G3PD method yields a satisfactory performance judged by visual inspection 
(see Figure \ref{figSegmentationResultDHBBOverview} for one example from each database)
and it outperforms the other methods on ten of twelve databases, see Table \ref{table:tableResults}. 
This demonstrates the robustness of the G3PD method for fingerprint segmentation.

\begin{table*}
   \begin{center}
  \begin{tabular}{|c|c|r@{.} l|r @{.} l|r @{.} l|r @{.} l|r @{.} l|r @{.} l|} \hline
	FVC	& DB	& \multicolumn{2}{c|}{GFB \cite{ShenKotKoo2001}} & \multicolumn{2}{c|}{HCR \cite{WuTulyakovGovindaraju2007}}	& \multicolumn{2}{c|}{MVC \cite{BazenGerez2001}}	& \multicolumn{2}{c|}{STFT \cite{ChikkerurCartwrightGovindaraju2007}}		& \multicolumn{2}{c|}{FDB \cite{ThaiHuckemannGottschlich2015}}	& \multicolumn{2}{c|}{G3PD}	
	\\
	\hline
	2000	& 1	& 13&26				& 11&15						& 10&01					& 16&70							& {\bfseries 5}&{\bfseries 51}			 	& 5&69 
		\\
		& 2	& 10&27				& 6&25						& 12&31					& 8&88							& {\bfseries3}&{\bfseries 55}			 	& 4&10
		\\
		& 3	& 10&63				& 7&80						& 7&45					& 6&44							& 2&86			 	& {\bfseries 2}&{\bfseries 68}
		\\
		& 4	& 5&17				& 3&23						& 9&74					& 7&19							& 2&31			 	& {\bfseries 2}&{\bfseries 06}
	\\
	\hline
	2002	& 1	& 5&07				& 3&71						& 4&59					& 5&49							& 2&39			 	& {\bfseries 1}&{\bfseries 72}
		\\
		& 2	& 7&76				& 5&72						& 4&32					& 6&27							& 2&91			 	& {\bfseries 2}&{\bfseries 83}
		\\
		& 3	& 9&60				& 4&71						& 5&29					& 5&13							& 3&35			 	& {\bfseries 3}&{\bfseries 27}
		\\
		& 4	& 7&67				& 6&85						& 6&12					& 7&70							& 4&49				& {\bfseries 3}&{\bfseries 63}
	\\
	\hline
	2004	& 1	& 5&00				& 2&26						& 2&22					& 2&65							& 1&40			 	& {\bfseries 0}&{\bfseries 88}
		\\
		& 2	& 11&18				& 7&54						& 8&06					& 9&89							& 4&90			 	& {\bfseries 4}&{\bfseries 62}
		\\
		& 3	& 8&37				& 4&96						& 3&42					& 9&35							& 3&14		 		& {\bfseries 2}&{\bfseries 77}
		\\
		& 4	& 5&96				& 5&15						& 4&58					& 5&18							& 2&79		 		& {\bfseries 2}&{\bfseries 53}
	\\
	\hline
	Avg.	& 	& 8&33				& 5&78						& 6&51					& 7&57							& 3&30				& {\bfseries 3}&{\bfseries 06}
	\\
	\hline 
  \end{tabular}
  \vspace*{8pt}
  \caption{ 
  Error rates (average percentage of misclassified pixels averaged over 800 test images per database)
  computed using the manually marked ground truth segmentation 
  and the estimated segmentation by these methods:
  a Gabor filter bank (GFB) response based method by Shen \textit{et al.} \cite{ShenKotKoo2001},
  a Harris corner response (HCR) based approach by Wu \textit{et al.} \cite{WuTulyakovGovindaraju2007},
  a method by Bazen and Gerez using local grey-level mean, variance and gradient coherence (MVC)
  as features \cite{BazenGerez2001},
  a method applying short time Fourier transforms (STFT) by Chikkerur \textit{et al.} \cite{ChikkerurCartwrightGovindaraju2007},
  the factorized directional bandpass (FDB) \cite{ThaiHuckemannGottschlich2015}
  and the proposed method based on the G3PD model.
  \label{table:tableResults}}
  \end{center}  
  \end{table*}

\section{Conclusions}
\label{sec:conclusions}

We have presented a global framework for the fingerprint segmentation problem 
which is to separate the foreground from background based on texture analysis. 
We have proposed the G3PD method for three-part decomposition of fingerprint images.
The texture pattern is analyzed under the variational approach 
considering sparsity and smoothness at the same time:
with the $\ell_1$-norm for sparsity 
and $\ell_1$-norm of curvelet coefficients for smoothness. 
The resulting texture image is binarised and postprocessed by morphology
to obtain the region of interest.

We have proposed a model for three-part decomposition
which takes the nature of the texture occurring in real fingerprint images into account.
Fingerprint images are characterized by a smooth, curved and oriented pattern 
which has a sparse representation in certain transform domains.

The G3PD method  
is somewhat similar in spirit to the FDB method \cite{ThaiHuckemannGottschlich2015}
which also takes into account the specific properties 
of fingerprint patterns. 
Frequencies occurring in real fingerprints 
are mostly located in a specific range in the Fourier domain 
and the corresponding texture is extracted by an elaborate bandpass filtering process
involving forward prediction, proximity operator and backward projection.
Similarly, the three-part decomposition can be regarded as 
lowpass, bandpass and highpass filtering of signals 
corresponding to $\boldsymbol u$, $\boldsymbol v$ and $\boldsymbol \epsilon$, 
respectively (see images (f-h) in Figure \ref{fig_Imagedecompose}).
This illustrates the connection between classical bandpass filtering
in the Fourier domain and the variational approach.

In conclusion, we have performed an extensive comparison of the G3PD method
with five state-of-the-art fingerprint segmentation algorithms
on a large benchmark with a variety of different challenges
and have found that the G3PD method outperforms its competitors
on ten out of twelve database in terms of segmentation accuracy.

We believe that this work paves the way for further research in areas such as latent fingerprint segmentation 
in which we deal additionally with other kinds of noise like large scale structure noise,
or to better deal with the few low-quality examples 
which still pose problems to the method.
We believe that further improvements can be achieved by combining the G3PD method 
with additional features, e.g. the texture image obtained by the FDB method.

\section*{Data Availability Statement}

Matlab Implementation of the G3PD Method for Fingerprint Segmentation\\ 
\url{http://dx.doi.org/10.6084/m9.figshare.1418020}\\\\
Benchmark for Fingerprint Segmentation Performance Evaluation\\ 
\url{http://dx.doi.org/10.6084/m9.figshare.1294209}\\\\
Matlab Implementation of the FDB Method for Fingerprint Segmentation\\
\url{http://dx.doi.org/10.6084/m9.figshare.1294210}\\\\
FVC databases\\
\url{http://bias.csr.unibo.it/fvc2000/}\\
\url{http://bias.csr.unibo.it/fvc2002/}\\
\url{http://bias.csr.unibo.it/fvc2004/}

\section*{Acknowledgements}

The authors would like to thank Axel Munk, Florian V\"ollering, and Benjamin Eltzner for their valuable comments.
C. Gottschlich gratefully acknowledges the support of the 
Felix-Bernstein-Institute for Mathematical Statistics in the Biosciences 
and the Niedersachsen Vorab of the Volkswagen Foundation.


\end{document}